\documentclass{article} 
\usepackage{iclr2025_conference,times}

\usepackage{amsmath,amsfonts,bm}

\def\eqref#1{equation~\ref{#1}}

\def\1{\bm{1}}

\DeclareMathAlphabet{\mathsfit}{\encodingdefault}{\sfdefault}{m}{sl}
\SetMathAlphabet{\mathsfit}{bold}{\encodingdefault}{\sfdefault}{bx}{n}

\newcommand{\Cov}{\mathrm{Cov}}

\usepackage{hyperref}
\usepackage{url}

\title{PRISM: PRIVACY-PRESERVING IMPROVED STOCHASTIC MASKING FOR FEDERATED GENERATIVE MODELS}

\iclrfinalcopy

\author{
    \textbf{Kyeongkook Seo}$^{1}$, \textbf{Dong-Jun Han}$^{2\:\ast}$, \textbf{Jaejun Yoo}$^{1}$\thanks{Co-corresponding author.} \\
    $^{1}$Ulsan National Institute of Science and Technology (UNIST) \\
    $^{2}$Department of Computer Science and Engineering, Yonsei University \\
    \texttt{\{kyeongkookseo, jaejun.yoo\}@unist.ac.kr, djh@yonsei.ac.kr}
}

\usepackage{kotex}
\usepackage{comment}
\usepackage{amssymb}
\usepackage{amsmath}
\usepackage{booktabs}
\usepackage{bbm}
\usepackage{pifont}
\usepackage{xcolor}
\usepackage{verbatim}
\usepackage{subcaption}
\usepackage{graphicx}
\usepackage{algorithm}
\newtheorem{theorem}{Theorem}
\newtheorem{definition}{Definition}
\usepackage{multirow}

\usepackage{wrapfig}
\usepackage{makecell}
\usepackage{algpseudocode}
\usepackage{xcolor, soul}

\sethlcolor{yellow}

\newcommand{\Fref}[1]{Figure \ref{#1}}
\newcommand{\Sref}[1]{Section \ref{#1}}
\newcommand{\Tref}[1]{Table \ref{#1}}
\newcommand{\Aref}[1]{Appendix \ref{#1}}
\newcommand{\cit}[1]{\citep{#1}]}

\newcommand{\prism}[1]{PRISM}
\newcommand{\prismalpha}[1]{PRISM-$\alpha$}
\newcommand{\prismdema}[1]{MADA}

\begin{document}

\maketitle
\begin{abstract}
Despite recent advancements in federated learning (FL), the integration of generative models into FL has been limited due to challenges such as high communication costs and unstable training in heterogeneous data environments. To address these issues, we propose \prism{}, a FL framework tailored for generative models that ensures (i) stable performance in heterogeneous data distributions and (ii) resource efficiency in terms of communication cost and final model size. The key of our method is to search for an optimal stochastic binary mask for a random network rather than updating the model weights, identifying a sparse subnetwork with high generative performance; \textit{i.e.}, a ``strong lottery ticket''. By communicating binary masks in a stochastic manner, \prism{} minimizes communication overhead. Combined with the utilization of maximum mean discrepancy (MMD) loss and a mask-aware dynamic moving average aggregation method (\prismdema{}) on the server side, \prism{}~facilitates stable and strong generative capabilities by mitigating local divergence in FL scenarios. Moreover, thanks to its sparsifying characteristic, \prism{}~yields an lightweight model without extra pruning or quantization, making it ideal for environments such as edge devices.
Experiments on MNIST, FMNIST, CelebA, and CIFAR10 demonstrate that \prism{}~outperforms existing methods, while maintaining privacy with minimal communication costs. \prism{} is the first to successfully generate images under challenging non-IID and privacy-preserving FL environments on complex datasets, where previous methods have struggled. Our code is available at \href{https://github.com/tjrudrnr2/PRISM_ICLR25}{PRISM}.

\end{abstract}
\section{Introduction}
\label{sec:introduction}

Recent generative models have demonstrated remarkable advancements in image quality and have been widely extended to various domains, including image-to-image translation~\cit{starganv2, ItI_diff}, layout generation~\cit{posterllama}, text-to-image generation~\cit{ldm, dalle2}, and video generation~\cit{stylegan-v, HVDM}. Achieving high-quality generation with current generative models demands increasingly large datasets, leading to concerns that  publicly available data will soon exhausted~\cit{data_scarce}. Leveraging the vast amount of data stored on edge devices becomes a potential solution, but this poses significant challenges: not only does the private nature of the data make centralized training impractical, but edge computing itself faces hurdles, including limited resources and prohibitive communication costs.

Federated learning (FL)~\cit{FL} is a promising paradigm tailored to this setup, enabling clients to collaboratively train a global model without sharing their local datasets with a third party. However, high communication costs, performance degradation due to data heterogeneity, and the need to preserve privacy remain significant challenges in FL. These challenges are further intensified in the context of generative models. Unlike classification tasks, generative tasks lack a well-defined objective function and focus on learning the data sample distribution, making the integration of FL and generative models even more difficult. A few recent works have made efforts to train generative models over distributed clients~\cit{mdgan, fedgan, iflgan, uagan, multiflgan}. These methods are generally built upon generative adversarial networks (GANs)~\cit{gan}, which have shown impressive results in the field of image generation. 
DP-FedAvgGAN~\cit{dpfedavggan}, GS-WGAN~\cit{gswgan}, and Private-FLGAN~\cit{private-flgan} apply differential privacy (DP)~\cit{dp, rdp} to mitigate the potential privacy risk in FL setups. However, existing works still face several challenges: 1) Previous approaches~\cit{gan_unstable_1, gan_unstable_2, gan_unstable_3} underperform, especially in non-IID (independent, identically distributed) data distribution scenarios with strong data heterogeneity across FL clients due to the notorious instability of GANs. 2) Performance evaluations are limited to relatively simple datasets such as MNIST, Fashion MNIST, and EMNIST. 3) They suffer from significant communication overhead during model exchanges between the server and clients. 

To overcome these challenges, we propose PRivacy-preserving Improved Stochastic-Masking for generative models (\prism{}), a new strategy for training generative models in FL settings with the following key features: \textbf{First},  
at the heart of \prism{}~is the strong lottery ticket (SLT) hypothesis~\cit{slt}, suggesting the existence of a highly effective subnetwork within a randomly initialized network. \prism{}~shifts the focus towards identifying an optimal global binary mask, rather than updating the weights directly. 
This approach enables each client to transfer the binary mask to the server instead of the full model, significantly reducing the overload in each communication round. Moreover, when training is finished, \prism{}~produces a lightweight final model, as each weight is already quantized, thanks to our initialization strategy. This feature provides significant advantages for resource-constrained edge devices.  
\textbf{Second}, \prism{} incorporates the maximum mean discrepancy (MMD) loss~\cit{mmd_1, mmd_2} during client-side updates, ensuring stable training for generative models. 
\textbf{Third}, a mask-aware dynamic moving average aggregation (\prismdema{}) is introduced to alleviate local model divergence. This allows \prism{} to maintain the previous global mask information and alleviate client drift under non-IID and DP-guaranteeing scenarios. 
By automatically adjusting the moving average parameter based on mask correlations, this approach requires neither a regularization term nor hyperparameter tuning. \textbf{Finally,} \prism{} offers a hybrid strategy that can flexibly trade-off between image quality and communication cost.
Taken together, these features enable \prism{} to consistently deliver robust performance in challenging non-IID and DP-guaranteeing FL settings, while maintaining minimal communication overhead.

Our experimental results reveal that \prism{}~sets a new standard in generative model performance, significantly outperforming GAN-based methods in both IID and non-IID scenarios. It acheives state-of-the-art image generation on complex datasets such as CelebA and CIFAR10, whereas previous methods were limited to simpler datasets like MNIST and FMNIST.  
This highlights \prism{}'s potential for scalable and resource-efficient generative model learning in distributed environments.  
 Overall, our main contributions can be summarized as follows:
\begin{itemize}
    \item We propose \prism{}, an effective FL framework that achieves state-of-the-art performance on various benchmark datasets. It is the first method to successfully generate images on complex datasets such as CelebA in FL scenarios that involves data heterogeneity and privacy preservation. 
    \item \prism{} offers an efficient solution for federated generative models with minimal communication overhead by incorporating SLT with a stochastic binary mask. Even more, in conjunction with the weight initialization strategy, the final model acquired from \prism{} becomes significantly lightweight, reducing to less than half the size of the initial model.
    \item We further enhance the stability of federated learning for generative models by introducing MMD loss and a mask-aware dynamic moving average aggregation method (\prismdema{}). 
\end{itemize}

To the best of our knowledge, this is the first work to address the challenges in communication efficiency, privacy, stability, and generation performance altogether for federated generative models. We provide new directions to this area based on several unique characteristics, including SLT with stochastic binary mask, MMD loss, mask-aware dynamic moving average aggregation strategy, and hybrid score/mask communications.
\vspace{-1mm}
\section{Related Work}
\label{sec:related_work}
\vspace{-1mm}

\textbf{Federated learning for classification models.}
FL has achieved a significant success in training a global model in a distributed setup, eliminating the necessity of sharing individual client’s local datasets with either the server or other clients. Research has been conducted on various aspects of FL, such as data heterogeneity~\cit{Fed_non-iid_1, Fed_non-iid_2}, communication efficiency~\cit{FedPM, Fedmask, fed_effi_1, fed_effi_2}, privacy~\cit{fed_dp}, with most focusing on image classification tasks. 
Related to our approach, FedPM~\cit{FedPM} and FedMask~\cit{Fedmask} adopted binary mask communication to reduce the communication costs in FL in classification tasks. 
FedMask~\cit{Fedmask} introduces binary mask communication, focusing on communication efficiency and personalization in decentralized environments, while FedPM~\cit{FedPM} utilizes stochastic masks to minimize uplink overhead and proposes a bayesian aggregation method to robustly manage scenarios with partial client participation. 
While the concept of incorporating SLT into FL paradigm has been studied in FedMask~\cit{Fedmask} and FedPM~\cit{FedPM} for \textit{classification tasks}, \prism{} stands as an independent strategy tailored for the training of \textit{generative models} across distributed clients:  \prism{} incorporates MMD loss for more robust performance compared to GAN-based approaches and introduces \prismdema{} to maintain a stable image generation performance in heterogeneous and DP-guaranteeing FL settings.

\medskip
\noindent \textbf{Federated learning for generative models.}
Generative models, like classification models, also offer potential benefits in various federated settings (See~\Aref{appendix:practical_scenarios}). Several recent works have aimed to incorporate generative models into distributed settings~\cit{mdgan, multiflgan, iflgan, uagan, fedgan, dpfedavggan, gswgan, private-flgan}. MD-GAN~\cit{mdgan} was the first attempt to apply generative models in the FL framework using GANs~\cit{gan}, extensively studied in image generation tasks. In MD-GAN, each client holds a discriminator, and the server aggregates the discriminator feedback from each client to train the global generator. To prevent overfitting of local discriminators, clients exchange discriminators, incurring additional communication costs. Multi-FLGAN~\cit{multiflgan} proposed all vs. all game approach by employing multiple generators and multiple discriminators and then selecting the most powerful network to enhance model performance. IFL-GAN~\cit{iflgan} improves both performance and stability by weighting each client's feedback based on the MMD between the images generated by the global model and the local generator. This approach maintains a balance between the generator and the discriminator, leading to Nash Equilibrium. Other works such as~\cit{uagan, fedgan} have also explored the utilization of GANs in FL. However, these works do not consider the challenge of privacy preservation in the context of FL and also suffer from resource issues during training and inference.

\noindent \textbf{Federated learning for generative models with privacy consideration.}
Only a few prior works have focused on the privacy preservation in federated generative models~\cit{dpfedavggan, gswgan}. DP-FedAvgGAN~\cit{dpfedavggan} introduces to combine federated generative models and differential privacy (DP)~\cit{dp, rdp} to ensure privacy preservation. GS-WGAN~\cit{gswgan} adopts Wasserstein GAN~\cit{wgan-gp} to bypass the cumbersome searching for an appropriate DP-value, leveraging the Lipschitz property. While these approaches have successfully integrated FL and generative models, they inherit drawbacks such as the notorious instability of GANs~\cit{gan_unstable_1, gan_unstable_2, gan_unstable_3} and significant performance drop under data heterogeneity. Moreover, all existing approaches suffer from huge communication and storage costs during and after training, respectively.

\vspace{-1mm}
\section{Background}
\label{sec:background}
\vspace{-1mm}

\subsection{Strong lottery tickets}
\label{subsec:slt}

Strong Lottery Ticket (SLT) hypothesis~\cit{slt, slt2, slt3} suggest the existence of a sparse subnetwork within an initially random network that achieves a superior performance. Edge-Popup (EP) algorithm~\cit{ep} is one of the most popular methods to discover SLT within the dense network, which introduces a scoring mechanism to select potentially important weights among the widespread initialized weight values. More specifically, given a randomly initialized dense network $W_{init}$, a learnable score $s$ is trained while keeping the weight values frozen. 
These scores are designed to encapsulate the importance of each weight for the objective function. As the scores get iteratively updated, the EP algorithm progressively shrinks the model by applying binary masks to weights with higher scores, indicating their potentials to be included in the winning lottery ticket. 
The obtained SLT can be expressed as $W=W_{init}\odot M$, where $M$ is the obtained binary mask and $\odot$ denotes element-wise multiplication. SLT has been primarily explored within the context of classification tasks, while Yeo \textit{et al.}~\cit{ep_mmd} have recently shown that SLT can also be found in generative models. 

\subsection{Differential privacy}
\label{subsec:dp}

Sharing each client's model or gradient   can potentially lead to a privacy risk. $(\epsilon, \delta)$-differential privacy (DP)~\cit{dp}, $(\alpha, \epsilon)$-R\'{e}nyi-Differential privacy (RDP)~\cit{rdp} are commonly employed when tackling the privacy concerns in FL. 
\begin{definition}[($\epsilon, \delta$)-Differential Privacy~\cit{dp}]
A randomized mechanism $\mathcal{M} : \mathcal{X} \rightarrow \mathcal{R}$ is $(\epsilon, \delta)$-differential privacy, if for any two adjacent datasets $\mathcal{D}$, $\mathcal{D'}$ and for any measurable sets $\mathcal{S}$:  $Pr[\mathcal{M}(D) \in \mathcal{S}] \leq e^{\epsilon} Pr[\mathcal{M}(D') \in \mathcal{S}] + \delta$.
 \end{definition}

\noindent The above definition is designed to limit the impact of individual data points by introducing randomness into $\mathcal{M}$. The Gaussian mechanism~\cit{noisy_subsample} offers differential privacy guarantees by injecting Gaussian noise $\mathcal{N}(0,\sigma^2I)$ to $\mathcal{M}$, where $\sigma^2=\frac{2ln(1.25/\delta)\Delta_2^2}{\epsilon^2}$ and $\Delta_2^2$ is L2 sensitivity.

\vspace{-1mm}

\section{Method}
\label{sec:method}
\vspace{-1mm}

\begin{figure}[t]
\vspace{-7mm}
  \centering
  \includegraphics[width=\textwidth]{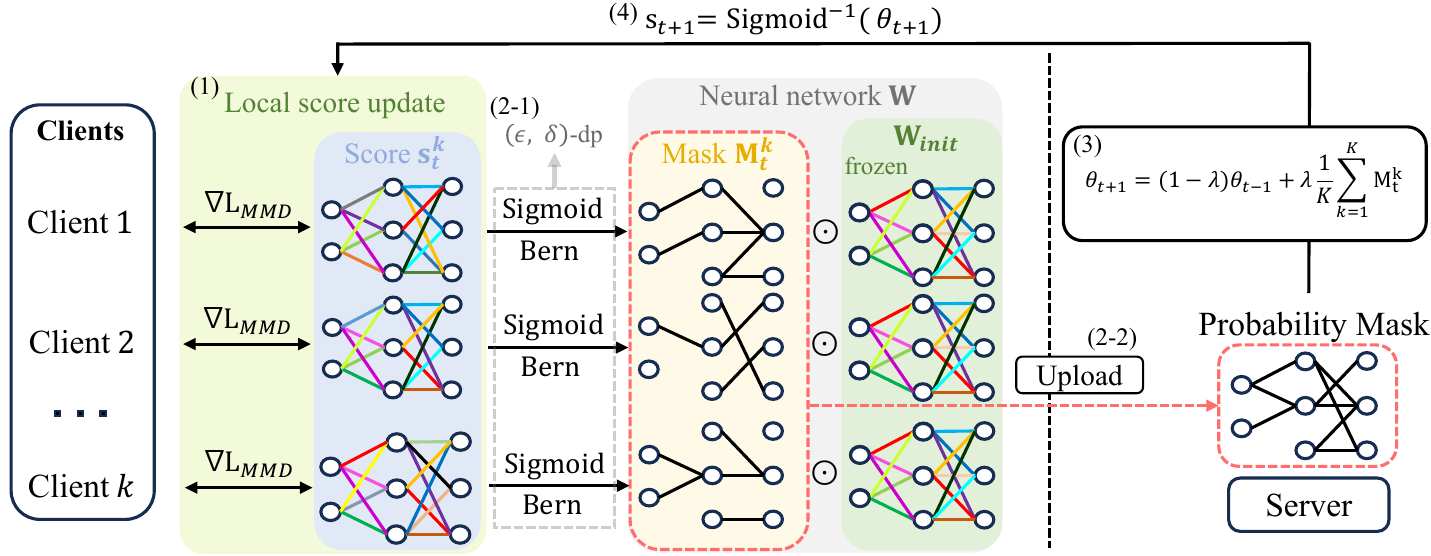}/

  \caption{\textbf{Overview of \prism{}.} \prism{} finds the supermask for generative models in a FL scenario. At every round $t$, each client $k$ updates a local score $s_t^k$ via MMD loss (Step 1) and generates the privacy-preserving binary mask $M_t^k$ (Step 2-1), which is sent to the server. The server aggregates the masks to obtain the global probability $\theta_{t+1}$ (Step 3), which is converted to a score $s_{t+1}$ and broadcasted to the clients for the next round (Step 4). The global probability $\theta_{t+1}$ is gradually updated based on mask correlation $\lambda$ between $M_t^g$ and $M_{t-1}^g$.
  }
  \label{fig:overview}
  \vspace{-2mm}
\end{figure}

We consider a FL setup with $K$ clients, where each client $k$ has its own local dataset $\mathcal{D}^k$. Starting from a randomly initialized model $W_{init}$, the clients aim to collaboratively obtain a global generative model $W^*$ that well-reflects all data samples in the system, i.e., in $\cup_{k=1}^K\mathcal{D}^k$. 

\smallskip
\noindent \textbf{Overview of our approach.} 
\Fref{fig:overview} shows the overview of our \prism{}. \prism{}~finds a subnetwork with strong generative performance from the randomly initialized generative model $W_{init}$. Rather than updating $W_{init}$, its focus is to find an optimal binary mask $M^*$ that has either 1 or 0 in its element and construct the final global model as $W^*=W_{init}\odot M^*$. At a high-level,  each client $k$ generates a binary mask $M_t^k$  based on its local dataset at every communication round $t$, which is aggregated at the server. After repeating the process for multiple rounds $t=1,2,\dots, T$, \prism{}~produces  the final supermask $M^*=M_T$. Our approach, which aims to find the SLT in a federated generative model setting, is fundamentally different from existing FedGAN methods.
In the following, we describe the detailed training procedure of \prism{}~along with its advantages.
\vspace{-1mm}

\subsection{\prism{}~: Privacy-preserving Improved Stochastic Masking}
\label{subsec:PRISM}
\vspace{-1mm}

\noindent \textbf{Local score updates with MMD loss.} Before training starts, the server randomly initializes the model $W_{init}$ and broadcasts it to all clients, which remains fixed throughout the training process. At the start of each round $t$, all clients download the score vector  $s_t$ from the server, representing the importance of each parameter in $W_{init}$. Intuitively, if the score value of a specific parameter is high, the corresponding weight is more likely to be included in the final SLT. \prism{}~allows each client $k$ to update the score vector $s_t$ based on its local dataset to obtain $s_t^k$, which is used to generate the local mask. 
In this local score update procedure, we leverage maximum mean discrepancy (MMD) loss~\cit{mmd_1, mmd_2}, providing stable convergence for training generative models~\cit{gmmn, mmdgan, demystifying_mmdgan, gfmn, ep, ep_mmd}. The MMD loss measures the distance between two distributions by comparing their respective mean embeddings in a reproducing kernel hilbert space (RKHS)~\cit{mmd_1, mmd_2}.
As in~\cit{gfmn, ep}, we take  VGGNet pretrained on ImageNet as a powerful characteristic kernel. Specifically, given the local dataset $\mathcal{D}^k=\{x_i^k\}_{i=1}^N$ of client $k$ and the fake image set $\mathcal{D}_{fake}^k=\{y_i^k\}_{i=1}^M$ produced by its own generator, the local objective function at each client $k$ is written as follows:
\begin{equation}
    \mathcal{L}_{MMD}^k=\Big\| \mathbb{E}_{x\sim \mathcal{D}^k}[\psi(x)] - \mathbb{E}_{y\sim \mathcal{D}_{fake}^k}[\psi(y)]\Big\|^2 + \Big\| \Cov(\psi(\mathcal{D}^k)) - \Cov(\psi(\mathcal{D}_{fake}^k))\Big\|^2,
\label{eq:mmd2}
\end{equation}
where $\psi(\cdot)$ is a function that maps each sample to the VGG embedding space. Each client aims to match the mean and covariance between real and fake samples after mapping them to the VGG embedding space using kernel $\psi(\cdot)$. 
Based on  Eq.~\ref{eq:mmd2}, each client locally updates the scores to minimize the MMD loss according to $s_t^k = s_t -\eta \nabla \mathcal{L}_{MMD}^k$. Here, we note that the VGG network is utilized only for computing the MMD loss and is discarded when training is finished.

\smallskip
\noindent \textbf{Binary mask generation and aggregation.} After the local score update process, each client $k$ maps the score $s_t^k$ to a probability value $\theta_t^k\in [0,1]$ as $\theta_t^k = Sigmoid(s_t^k)$, where $Sigmoid(\cdot)$ is the sigmoid function. The obtained $\theta_t^k$  is then used as the parameter of the Bernoulli distribution to generate the stochastic binary mask $M_t^k$, according to $M_t^k\sim Bern(\theta_t^k)$. Each client $k$ uploads only this binary mask $M_t^k$ to the server, significantly reducing the communication overhead. 
At the server side, the received masks are aggregated to estimate the global Bernoulli parameter as $\theta_{t+1} = \frac{1}{K}\sum_{k=1}^KM_t^k$, which can be interpreted as the probabilistic score reflecting the importance of the overall client's weights. $\theta_{t+1}$ is then converted to the score through the inverse of the sigmoid function according to $s_{t+1}=Sigmoid^{-1}(\theta_{t+1})$, which is broadcasted to the clients at the beginning of the next round. We provide the training details in \Aref{appendix:training_details}.
 
\smallskip
\noindent \textbf{Model initialization for storage efficiency.} 
When training is finished after $T$ rounds of FL, $\theta_T$ is obtained at the server. The supermask is then generated following $M^*\sim Bern(\theta_T)$, which is used to obtain the final global model as $W^*=W_{init}\odot M^*$. 
This final model $W^*$ can be stored efficiently even in resource-constrained edge devices, thanks to the model initialization strategy. When initializing $W_{init}$ in \prism{}, we employ the standard deviation of Kaiming Normal distribution~\cit{kaiming_normal}, where the weight value in layer $l$ is sampled from the set $\sigma_l \sim \{-\sqrt{2/n_{l-1}}, \sqrt{2/n_{l-1}} \}$. Hence, by storing the scaling factor $\sqrt{2/n_{n_{l-1}}}$, each parameter in the initial model $W_{init}$ is already quantized to a 1-bit value (See~\Aref{appendix:quantization}). This makes the final model exceptionally lightweight without extra pruning of quantization, which will be also showed via comparison in~\Sref{subsec:exp_final_model}.
\vspace{-1mm}

\subsection{Privacy}
\vspace{-1mm}

To consider the situation of potential privacy treats, we incorporate $(\epsilon, \delta)$-differential privacy (DP)~\cit{dp} into our framework. We provide a more detailed description related to the privacy preservation of \prism{} and a specific scenario where \prism{} can achieve additional privacy benefits in \Aref{appendix:dp} and \Aref{appendix:privacy}, respectively.

\vspace{-1mm}

\subsection{Mask-Aware Dynamic moving average aggregation}
\vspace{-1mm}

Data heterogeneity and privacy preservation pose significant challenges in accurately estimating the correct update direction. Both the previous and current global models contain valuable information about whether the local models are diverging~\cite{scaffold, fedprox, fedalign}. To address this, we propose a mask-aware dynamic moving average aggregation (\prismdema{}) that leverages information from previous aggregation rounds in a mask-aware manner. To the best of our knowledge,~\prismdema{} is proposed for the first time, introducing the mask-aware similarity to determine the moving average ratio on the server side. Specifically, when a client mask $M_t^k$ deviates significantly from the global model, the newly updated global mask $M_t^g$ will also differ considerably from the previous round's global mask $M_{t-1}^g$. However, such updates tend to favor the dominant clients. To mitigate this bias, the server calculates the \textit{mask correlation} $\lambda$, which measures the similarity between the current and previous global masks ($M_t^g$ and $M_{t-1}^g$, respectively). The server then interpolates between the two global masks using $\lambda$ to adjust how much of the current mask should be incorporated. While various metrics can be used to compute the distance between masks, we employ the Hamming distance. We observe that using alternative distance metrics like cosine similarity yields similar performance (See \Aref{appendix:cos}). 
The server-side aggregation process is defined as follows:
\begin{equation}
\theta_{t+1}=(1-\lambda)\theta_{t-1} + \lambda \frac{1}{K}\sum_{k=1}^{K}M_t^k, \quad \lambda:=dist(M_{t-1}^g,M_{t}^g),
\label{eq:ema_aggregation}
\end{equation}
This prevents excessive deviation by interpolating the aggregated mask with the previous Bernoulli parameter. As the global rounds progress, $\lambda$ gradually decreases, promoting stable convergence.

\vspace{-1mm}
\subsection{Trading-off between performance and communication cost}
\label{subsec:noniid w/o dp}
\vspace{-1mm}
While \prism{} optimizes for minimal communication overhead and delivers satisfactory performance, it may not always meet the demand for higher-quality image generation. 
To address this, we introduces a flexible solution, \prism{}$^{\ast}$ (See~\Aref{appendix:hybrid}).

\vspace{-2mm}

\section{Experiments}
\label{sec:experiments}
\vspace{-2mm}

In this section, we validate the effectiveness of \prism{}~on MNIST, FMNIST, CelebA, and CIFAR10 datasets. The training set of each dataset is distributed across 10 clients following either IID or non-IID data distributions\footnote{See~\Aref{appendix:crossdevice} for additional results where a large number of clients and partial of clients participate.}. We provide the detailed partitioning strategies for IID and non-IID simulation in~\Aref{appendix:NonIID_detail}.
In addition, for a fair comparison, we set $(9.8, 10^{-5})$-DP for all methods.

\smallskip
\noindent \textbf{Baselines.} We compare our method with several previous approaches for federated generative models under both \textit{privacy-preserving} (with DP) and \textit{privacy-free} (without DP) scenarios:  In the privacy-preserving scenario, we consider DP-FedAvgGAN~\cit{dpfedavggan} and GS-WGAN~\cit{gswgan} while  in  the privacy-free case, we adopt MD-GAN~\cit{mdgan} and Multi-FLGAN~\cit{multiflgan}.

In the case of Multi-FLGAN, the number of sync servers increases the communication cost quadratically, so we consider a 2$\times$2 multi generator and discriminator setup. 

\smallskip
\noindent \textbf{Performance metrics.} We evaluate the generative performance of each scheme using the commonly adopted metrics, including Fr\'{e}chet Inception Distance (FID)~\cit{fid}, Precision \& Recall~\cit{p&r}, Density \& Coverage~\cit{d&c}. We further demonstrate the efficiency of \prism{}~by comparing the uplink cost (MB) (See \Aref{appendix:discussion_cost}) at each FL round, the storage (MB) for the final models, and FLOPS (See~\Aref{appendix:FLOPS}).

\vspace{-1mm}

\subsection{IID case}
\label{subsec:iid}
\vspace{-1mm}

In this subsection, we examine an IID scenario where the training set of each dataset is uniformly distributed among clients. We compare various evaluation metrics under $(\epsilon, \delta)$-DP guaranteed setting (\Tref{tab:iid_dp}). Notably, \prism{}~outperforms current GAN-based models by a large margin. 

\begin{figure}[t]
\vspace{-3mm}
  \centering
  \includegraphics[width=\textwidth]{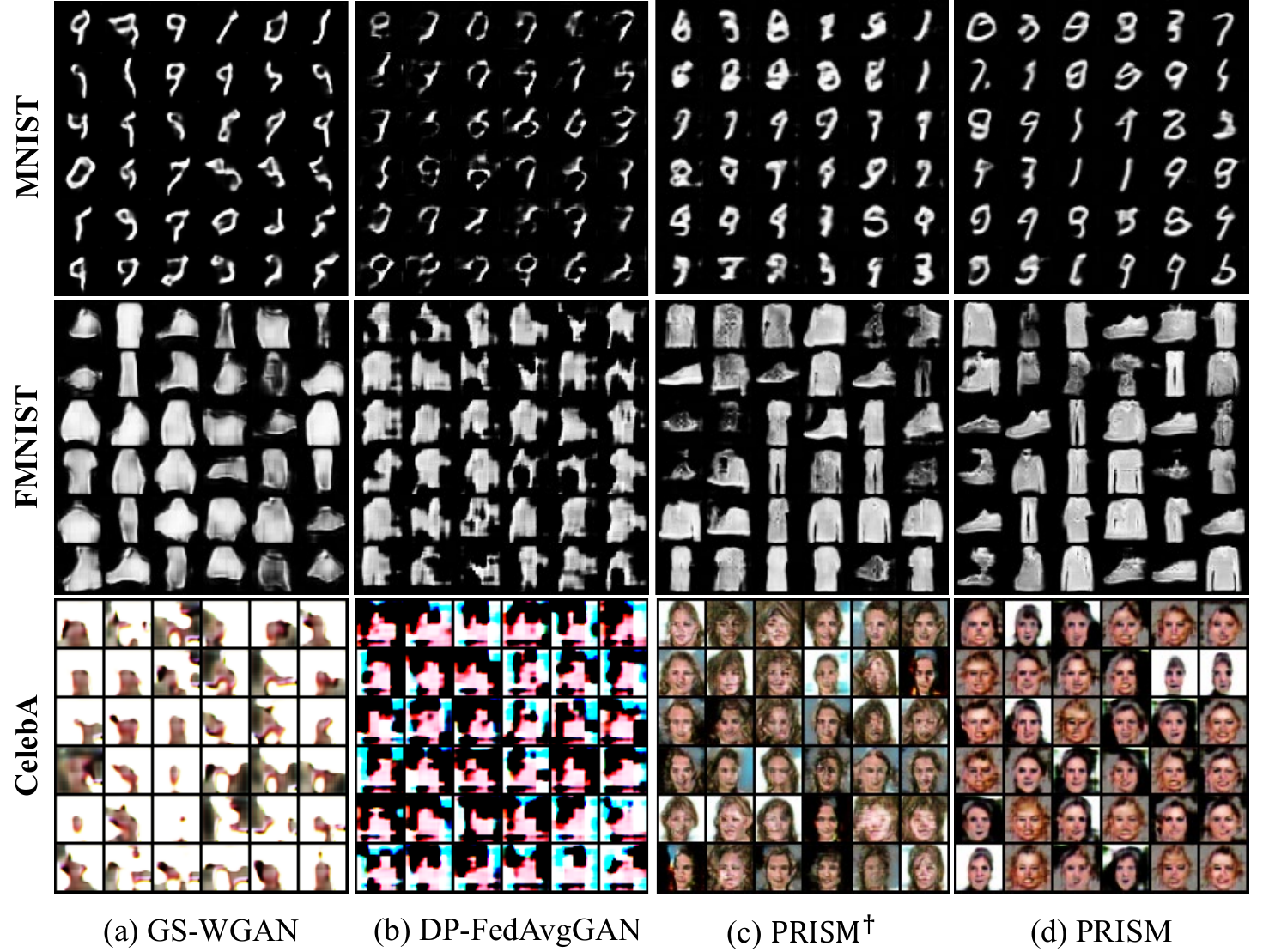}

  \caption{\textbf{Qualitative results in IID scenario with a privacy budget $(\epsilon, \delta)=(9.8, 10^{-5})$.} We compare generated images from the models in \Tref{tab:iid_dp} on MNIST, FMNIST, and CelebA. $\dagger$ indicates that \prismdema{} is removed.
  }
  \label{fig:iid_dp}
  \vspace{-5.5mm}
\end{figure}

\Fref{fig:iid_dp} reveals that existing privacy-preserving methods often produce distorted images, particularly evident in CelebA, while our method tends to generate high-quality results. The above results support that \prism{} can effectively find a SLT, achieving performance gains despite charging 48\% less communication costs per round. It is worth noting that \prism{} can further reduce the cost by applying extra techniques such as universal coding. Additionally, when comparing \prism{} to \prism{}$^{\dagger}$ (MADA removed), \prism{} demonstrates a significant improvement in overall performance.

\begin{table}[t!]
\vspace{-7mm}
\caption{\textbf{Quantitative comparison in IID scenario with a privacy budget $(\epsilon, \delta)=(9.8, 10^{-5})$.} We compare FID, P\&R, D\&C, communication cost, and storage. Communication cost is the number of bytes exchanged between clients and server. $\dagger$ indicates that \prismdema{} is removed.
}
\vspace{-2mm}
\begin{center}
\resizebox{0.9\textwidth}{!}{
\begin{tabular}{cccccc}
\toprule
\begin{tabular}[c]{@{}c@{}}\textbf{Method}\\ \footnotesize(comm.cost)\end{tabular} & \multirow{1}{*}{\textbf{Metric}} & \multirow{1}{*}{\textbf{MNIST}} & \multirow{1}{*}{\textbf{FMNIST}} & \multirow{1}{*}{\textbf{CelebA}} & \multirow{1}{*}{\textbf{Storage}} \\
\midrule
\multirow{3}{*}{\begin{tabular}[c]{@{}c@{}} GS-WGAN \\ \footnotesize(15MB) \end{tabular}}
    & \textbf{FID$\downarrow$} & 71.1016 & 119.2589 & 230.7874 & \multirow{3}{*}{\begin{tabular}[c]{@{}c@{}} 15MB \end{tabular}} \\
    & \textbf{P\&R $\uparrow$} & 0.0975 / 0.1505 & 0.3694 / 0.0015 & 0.7951 / 0.0 &  \\
    & \textbf{D\&C $\uparrow$} & 0.0257 / 0.0367 & 0.1264 / 0.0347 & 0.165 / 0.0021 &  \\ 
\midrule
\multirow{3}{*}{\begin{tabular}[c]{@{}c@{}} DP-FedAvgGAN \\ \footnotesize(14MB) \end{tabular}}
    & \textbf{FID $\downarrow$} & 111.0855 & 118.5067 & 221.34 & \multirow{3}{*}{\begin{tabular}[c]{@{}c@{}} 14MB \end{tabular}} \\
    & \textbf{P\&R $\uparrow$} & 0.2586 / 0.0047 & 0.5318 / 0.0163 & 0.1008 / 0.0 &  \\
    & \textbf{D\&C $\uparrow$} & 0.0803 / 0.0141 & 0.2028 / 0.0341 & 0.0211 / 0.0013 &  \\
\midrule
\multirow{3}{*}{\begin{tabular}[c]{@{}c@{}} \prism{}$^{\dagger}$ \\ \footnotesize(5.75MB) \end{tabular}}
    & \textbf{FID $\downarrow$} & 48.5636 & 54.722 & 57.0573 & \multirow{3}{*}{\begin{tabular}[c]{@{}c@{}} 7.25MB \end{tabular}} \\
    & \textbf{P\&R $\uparrow$} & 0.3343 / 0.4265 & 0.5836 / 0.1574 & 0.4998 / \textbf{0.1221} &  \\
    & \textbf{D\&C $\uparrow$} & 0.1211 / 0.1151 & 0.2156 / 0.2432 & 0.2572 / 0.2189 &  \\
\midrule
\multirow{3}{*}{\begin{tabular}[c]{@{}c@{}} \prism{} \\ \footnotesize(5.75MB) \end{tabular}}
    & \textbf{FID $\downarrow$} & \textbf{27.3017} & \textbf{46.1652} & \textbf{48.9983} & \multirow{3}{*}{\begin{tabular}[c]{@{}c@{}} 7.25MB \end{tabular}} \\
    & \textbf{P\&R $\uparrow$} & \textbf{0.4377} / \textbf{0.5576} & \textbf{0.6355} / \textbf{0.211} & \textbf{0.6435} / 0.076 &  \\
    & \textbf{D\&C $\uparrow$} & \textbf{0.1738} / \textbf{0.1982} & \textbf{0.4002} / \textbf{0.2971} & \textbf{0.4089} / \textbf{0.2415} &  \\
\bottomrule
\end{tabular}
}
\label{tab:iid_dp}
\end{center}
\end{table}

\begin{table}[t!]
\vspace{-7mm}
\caption{\textbf{Quantitative comparison in non-IID scenario with $(\epsilon, \delta)=(9.8, 10^{-5})$.} We compare FID, P\&R, D\&C, communication cost, and storage. Communication cost refers to the number of bytes exchanged between clients and server.
$\dagger$ indicates that MADA is removed.
}
\vspace{-2mm}
\begin{center}
\resizebox{0.9\textwidth}{!}{
\begin{tabular}{cccccc}
\toprule
\begin{tabular}[c]{@{}c@{}}\textbf{Method}\\ \footnotesize(comm.cost)\end{tabular} & \multirow{1}{*}{\textbf{Metric}} & \multirow{1}{*}{\textbf{MNIST}} & \multirow{1}{*}{\textbf{FMNIST}} & \multirow{1}{*}{\textbf{CelebA}} & \multirow{1}{*}{\textbf{Storage}} \\
\midrule
\multirow{3}{*}{\begin{tabular}[c]{@{}c@{}}
    GS-WGAN \\ \footnotesize(15MB) \end{tabular}}
                        & \textbf{FID $\downarrow$} & 338.6659 & 131.6166 &  228.9705 & \multirow{3}{*}{\begin{tabular}[c]{@{}c@{}} 15MB \end{tabular}} \\
                        & \textbf{P\&R $\uparrow$} & 0.0 / 0.0 & 0.4186 / 0.0001 & 0.1363 / 0.0 &   \\
                        & \textbf{D\&C $\uparrow$} & 0.0 / 0.0 & 0.1569 / 0.0297 & 0.0307 / 0.0025 &  \\
\midrule
\multirow{3}{*}{\begin{tabular}[c]{@{}c@{}}
    DP-FedAvgGAN \\ \footnotesize(14MB) \end{tabular}}
                        & \textbf{FID $\downarrow$} & 153.9325 & 146.632 & 222.8257 & \multirow{3}{*}{\begin{tabular}[c]{@{}c@{}} 14MB \end{tabular}} \\
                        & \textbf{P\&R $\uparrow$} & 0.4371 / 0.0336 & \textbf{0.7207} / 0.0043 & 0.2331 / 0.0004 &  \\
                        & \textbf{D\&C $\uparrow$} & 0.1049 / 0.004 & 0.2589 / 0.0164 & 0.0668 / 0.0016 &  \\
\midrule
\multirow{3}{*}{\begin{tabular}[c]{@{}c@{}}
  \prism{}$^{\dagger}$ \\ \footnotesize(5.75MB) \end{tabular}} 
                        & \textbf{FID $\downarrow$} & 49.6273 & 83.0481 & 59.4877 & \multirow{3}{*}{\begin{tabular}[c]{@{}c@{}} 7.25MB \end{tabular}}  \\
                        & \textbf{P\&R $\uparrow$} & 0.3283 / 0.3844 & 0.4513 / 0.0775 & 0.4789 / \textbf{0.0898} & \\
                        & \textbf{D\&C $\uparrow$} & 0.1101 / 0.1022 & 0.2355 / 0.1428 & 0.2392 / 0.2058 & \\
\midrule
\multirow{3}{*}{\begin{tabular}[c]{@{}c@{}} \prism{} \\ \footnotesize(5.75MB) \end{tabular}}
    & \textbf{FID $\downarrow$} & \textbf{34.2038} & \textbf{67.1648} & \textbf{39.7997} & \multirow{3}{*}{\begin{tabular}[c]{@{}c@{}} 7.25MB \end{tabular}} \\
    & \textbf{P\&R $\uparrow$} & \textbf{0.4386} / \textbf{0.4236} & 0.4967 / \textbf{0.1231} & \textbf{0.6294} / 0.0713 &  \\
    & \textbf{D\&C $\uparrow$} & \textbf{0.1734} / \textbf{0.1597} & \textbf{0.2748} / \textbf{0.1681} & \textbf{0.4565} / \textbf{0.2967} &  \\
\bottomrule
\end{tabular}
}
\label{tab:noniid_dp}
\end{center}
\end{table}

\begin{figure}[h!]
   \vspace{-5mm}
  \centering
  \includegraphics[width=0.98\textwidth]{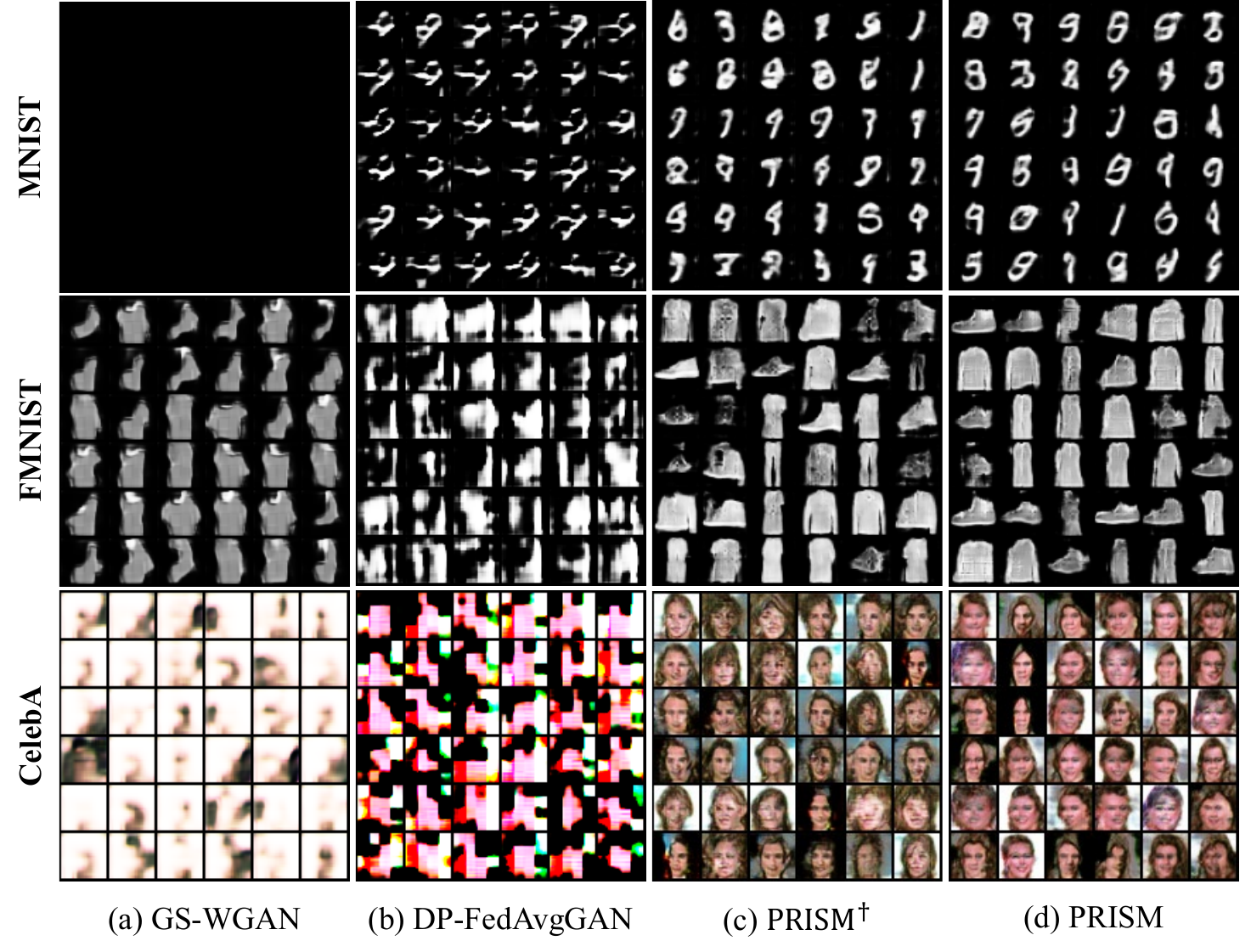}
      \vspace{-2mm}
  \caption{\textbf{Qualitative results in Non-IID scenario with a privacy budget $(\epsilon, \delta)=(9.8, 10^{-5})$.} We compare generated images from the models in \Tref{tab:noniid_dp} on MNIST, FMNIST, and CelebA. $\dagger$ indicates that \prismdema{} is removed.
  }
  \label{fig:noniid_dp}
   \vspace{-7mm}
\end{figure}

\begin{figure}[t!]
   \vspace{-7mm}
  \centering
  \includegraphics[width=\textwidth]{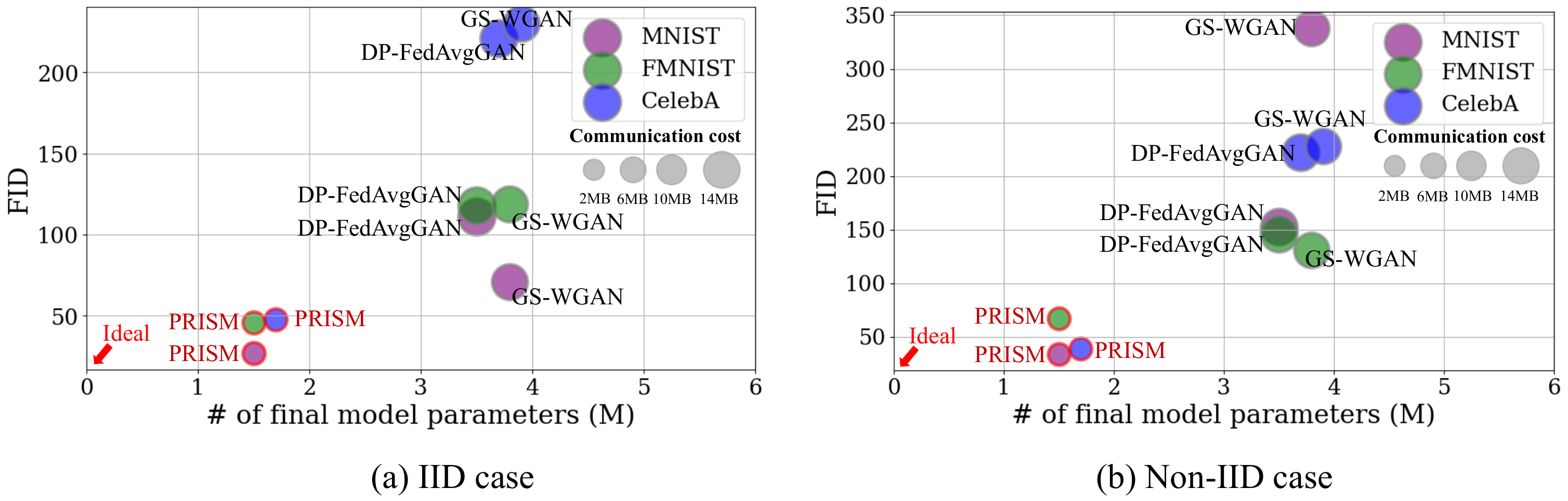}

   \vspace{-3mm}
  \caption{\textbf{The performance of baselines and our PRISM with privacy budget $(\epsilon, \delta)=(9.8, 10^{-5})$.} X-axis represents the number of parameters of final generator, while Y-axis represents FID. The diameter of each circle denotes the required communication cost at every round. The ideal case is the bottom-left corner.}
  \label{fig:Communication_costs}
\end{figure}

\begin{table}[t]
\vspace{-2mm}
\caption{\textbf{Quantitative comparison in non-IID scenario without DP.} We compare FID, P\&R, D\&C. Communication cost refers to the number of bytes exchanged between clients and server. $\dagger$ indicates that \prismdema{} is removed. We set $\alpha=80$ for \prism{}$^{\ast}$.
}
\vspace{-2mm}
\begin{center}
\resizebox{\textwidth}{!}{
\begin{tabular}{ccccccc}
\toprule
\begin{tabular}[c]{@{}c@{}} \textbf{Method} \\ \footnotesize(comm.cost) \end{tabular} & \multirow{1}{*}{\textbf{Metric}} & \multirow{1}{*}{\textbf{MNIST}} & \multirow{1}{*}{\textbf{FMNIST}} & \multirow{1}{*}{\textbf{CelebA}} & \multirow{1}{*}{\textbf{CIFAR10}} & \multirow{1}{*}{\textbf{Storage}} \\
\midrule
\multirow{3}{*}{\begin{tabular}[c]{@{}c@{}}
  MD-GAN \\ \footnotesize(14MB) \end{tabular}}  
                        & \textbf{FID $\downarrow$} & 37.7971 & 55.5094 & 18.907 & 52.7159 & \multirow{3}{*}{\begin{tabular}[c]{@{}c@{}} 14MB \end{tabular}} \\
                        & \textbf{P\&R $\uparrow$} & 0.3366 / 0.5435 & 0.5635 / 0.05 & 0.7612 / \textbf{0.6425} & \textbf{0.827} / 0.1968 \\
                        & \textbf{D\&C $\uparrow$} & 0.1192 / 0.1405 & 0.3145 / 0.2033 & 0.7238 / 0.4267 & \textbf{1.2201} / 0.3829 \\
\midrule
\multirow{3}{*}{\begin{tabular}[c]{@{}c@{}}
  Multi-FLGAN \\ \footnotesize(52MB) \end{tabular}}  
                        & \textbf{FID $\downarrow$} & 32.1014 & 125.9276 & 314.8386 & 163.0540 & \multirow{3}{*}{\begin{tabular}[c]{@{}c@{}} 14MB \end{tabular}} \\
                        & \textbf{P\&R $\uparrow$} & 0.5659 / 0.3353 & 0.4781 / 0.0035 & 0.0 / 0.0 & 0.9345 / 0.0 \\
                        & \textbf{D\&C $\uparrow$} & 0.3171 / 0.2709 & 0.24 / 0.0595 & 0.0 / 0.0 & 0.3638 / 0.0668 \\
\midrule
\multirow{3}{*}{\begin{tabular}[c]{@{}c@{}}
  \prism{}$^{\dagger}$ \\ \footnotesize(5.75MB) \end{tabular}} 
                        & \textbf{FID $\downarrow$} & 15.2329 & 35.1448 & 24.2591 & 68.4238 & \multirow{3}{*}{\begin{tabular}[c]{@{}c@{}} 7.25MB \end{tabular}} \\
                        & \textbf{P\&R $\uparrow$} & 0.7128 / 0.5289 & 0.7239 / 0.1049 & \textbf{0.7988} / 0.1868 & 0.65 / 0.1732 \\
                        & \textbf{D\&C $\uparrow$} & 0.5106 / 0.4851 & 0.645 / 0.3768 & \textbf{1.0746} / 0.5809 & 0.5575 / 0.3031 \\
\midrule
\multirow{3}{*}{\begin{tabular}[c]{@{}c@{}}
  \prism{} \\ \footnotesize(5.75MB) \end{tabular}} 
                        & \textbf{FID $\downarrow$} & 9.698 & 32.7517 & 21.8567 & 61.1198 & \multirow{3}{*}{\begin{tabular}[c]{@{}c@{}} 7.25MB \end{tabular}} \\
                        & \textbf{P\&R $\uparrow$} & 0.7665 / 0.6253 & \textbf{0.7614} / 0.1281 & 0.7835 / 0.1615 & 0.5924 / 0.2323 \\
                        & \textbf{D\&C $\uparrow$} & \textbf{0.6088} / 0.6003 & \textbf{0.8361} / \textbf{0.4383} & 1.047 / 0.6079 & 0.4334 / 0.3171 \\
\midrule
\multirow{3}{*}{\begin{tabular}[c]{@{}c@{}}
  \prism{}$^{\ast}$ \\ \footnotesize(15MB) \end{tabular}} 
                        & \textbf{FID $\downarrow$} & \textbf{6.9568} &  \textbf{29.0081} & \textbf{13.0209} & \textbf{35.5326} & \multirow{3}{*}{\begin{tabular}[c]{@{}c@{}} 7.25MB \end{tabular}} \\
                        & \textbf{P\&R $\uparrow$} & \textbf{0.7717} / \textbf{0.7992} & 0.697 / \textbf{0.1572} & 0.7893 / 0.392 & 0.6662 / \textbf{0.3642} \\
                        & \textbf{D\&C $\uparrow$} & 0.6082 / \textbf{0.6499} & 0.7002 / 0.4056 & 1.072 / \textbf{0.7396} & 0.5764 / \textbf{0.4481} \\
\bottomrule
\end{tabular}
}
\label{tab:noniid}
\end{center}
\vspace{-4mm}
\end{table}

\vspace{-2mm}

\subsection{Non-IID case}
\label{subsec:noniid}
\vspace{-1mm}

We investigate a more practical yet challenging non-IID scenario, where clients exhibit diverse local data distributions, posing a significant challenge to train generative models. To establish such an environment, we split the entire trainset based on shards-partitioning strategy\footnote{We also provide the results of Dirichlet-partitioned Non-IID in~\Aref{appendix:NonIID_detail}.}.
\Tref{tab:noniid_dp} presents a quantitative comparison of baselines and our methods when DP is guaranteed. We observe that~\prism{} achieves SOTA performance under the non-IID scenario, even with complex CelebA datasets\footnote{Since existing works on the field of federated generative models struggles with MNIST-level generation, we refer CelebA and CIFAR10 as complex dataset.}.
\Fref{fig:noniid_dp} illustrates that despite the heterogeneity of the data, our methods successfully generate high quality images, while traditional methods exhibit subpar quality. Additionally, \Fref{fig:Communication_costs} shows the overview of FID scores, final model parameters, and communication costs for each method.


\begin{figure}[t!]
   \vspace{-7mm}
  \centering
  \includegraphics[width=0.98\textwidth]{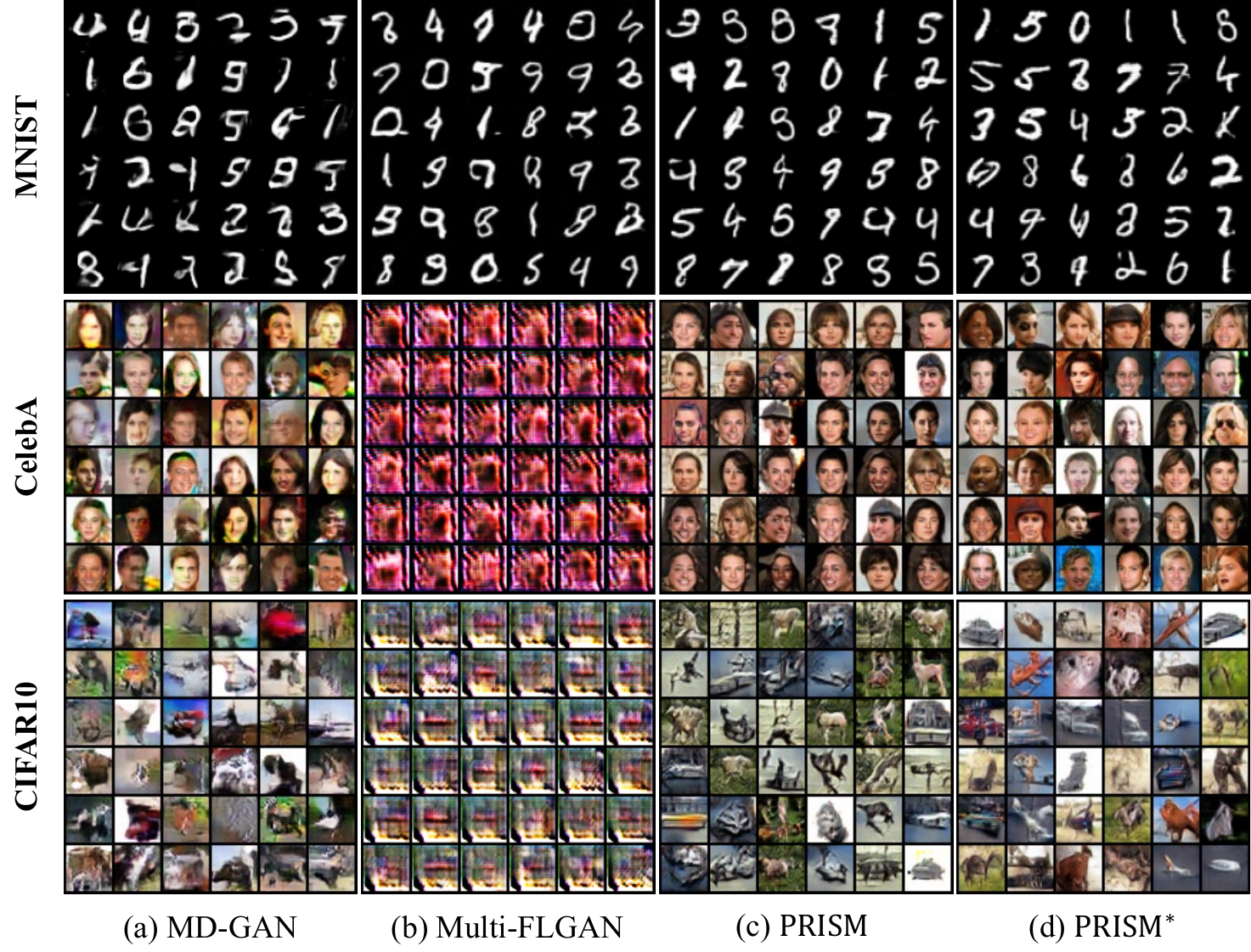}
  \vspace{-2mm}
  \caption{\textbf{Qualitative results in non-IID scenario without considering privacy budget.} Generated images from the models in \Tref{tab:noniid} on MNIST, FMNIST, CelebA, and CIFAR10. Here, we set $\alpha=80$ for \prism{}$^{\ast}$.} 
  \label{fig:noniid}
   \vspace{-3mm}
\end{figure}
\vspace{-1.5mm}
\subsection{Performance without differential privacy}
\label{subsec:noniid w/o dp}
\vspace{-1.5mm}

We further explore the performance of \prism{} without applying differential privacy. Quantitative and qualitative comparisons with MD-GAN~\cit{mdgan} and Multi-FLGAN~\cit{multiflgan}, the current state-of-the-art under this condition, are shown in  Table~\ref{tab:noniid} and Figure~\ref{fig:noniid}. \prism{} not only matches but also occasionally surpasses the performance of MD-GAN and Multi-FLGAN, all while significantly reducing communication overhead. Again, the fact that \prism{} outperforms \prism{}$^{\dagger}$ clearly demonstrates the effectiveness of MADA. Here, Multi-FLGAN takes 10 times more communication cost than \prism{} due to multi GAN strategy. However, in FL setups, users may want to trade-off between the communication cost and generative performance. As shown in~\Tref{tab:noniid} and~\Fref{fig:noniid}, \prism{}$^{\ast}$ outperforms across all benchmarks, including CIFAR10, while maintaining a communication cost comparable to MD-GAN and far lower than Multi-FLGAN. Further details on~\prism{}$^{\ast}$ can be found in~\Aref{appendix:hybrid}. \vspace{-1mm}

\subsection{Effect of dynamic moving average aggregation}
\label{subsec:ablation_DEMA}
   \vspace{-1mm}

\begin{figure}[t]
  \centering
  \includegraphics[width=\textwidth]{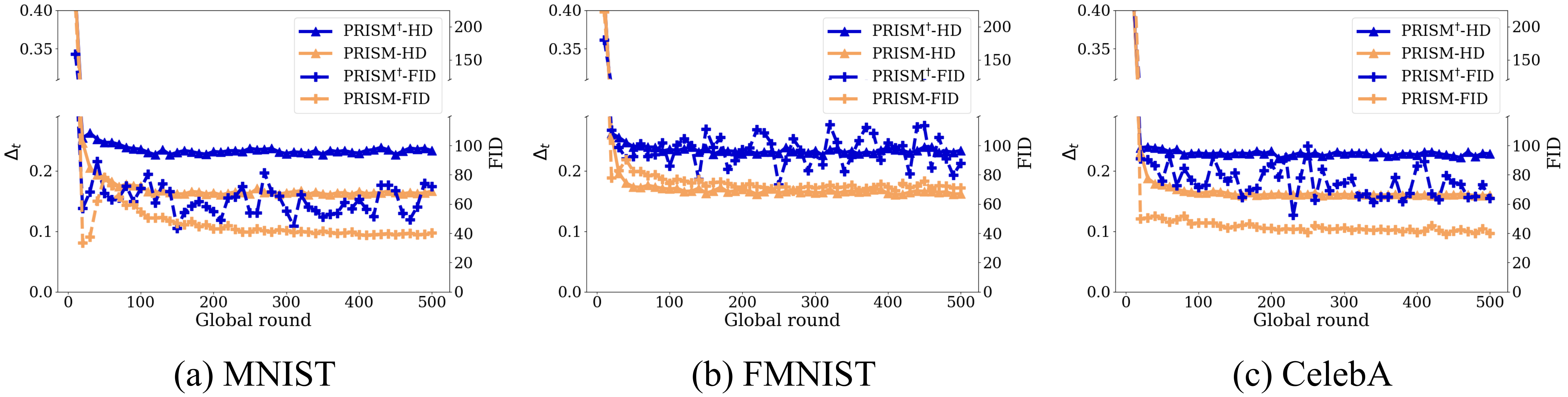}
  \vspace{-5mm}
  \caption{\textit{Local divergence} $\Delta_t$ and FID values in the non-IID and DP-considering scenario.}
  \label{fig:local-divergence}
    \vspace{-4mm}
\end{figure}
In this section, we empirically demonstrate the effectiveness of dynamic moving average aggregation using the MNIST dataset. \Fref{fig:local-divergence} visualizes local model updates over communication rounds $t$. At each global round $t$, clients receive the aggregated global mask $M_{t-1}^g$ and continue local training for several epochs. Afterward, the trained local mask $M_t^k$ is uploaded for the next communication round. We introduce the \textit{local divergence} metric $\Delta_t:=hd(M_t^g, M_t^k)$, which is defined as the Hamming distance between the received mask and the trained local mask to track the discrepancy of local model updates. For simplicity, we visualize the results for the first client, but similar trends were observed across other clients.  
In~\Fref{fig:local-divergence}, \prism{} exhibits impressive FID scores and reduced local updates than \prism{}$^{\dagger}$, clearly demonstrating that MADA not only restricts the local model divergence but also achieves significant performance gain across various challenging FL settings. This is accomplished by automatically obtaining $\lambda$ based on the current mask, without requiring additional regularization terms or hyperparameter tuning\footnote{The convergence of $\lambda$ in MADA and its relationship with client drift are discussed in \Aref{appendix:Analysis_MADA}.}.

\vspace{-1mm}
\subsection{Resource efficiency at inference time}
\label{subsec:exp_final_model}
\vspace{-1mm}
The final model sizes of the baselines and our method are reported in \Tref{tab:iid_dp}, \Tref{tab:noniid_dp}, and \Tref{tab:noniid}. 
Note that in addition to our method, applying various lossless compression techniques (\textit{e.g.,} arithmetic coding~\cit{arithmetic_coding}) can further reduce the required resources of \prism{}.
\vspace{-3mm}
\section{Conclusion}
\label{sec:conclusion}
\vspace{-3mm}
While image generation has emerged as promising area in deep learning, the fusion of FL with generative models remains relatively unexplored. In this paper, we proposed \prism{}, an efficient and stable federated generative framework that capitalizes on stochastic binary masks and MMD loss. 
To further enhance stability under non-IID and privacy-preserving scenario, we introduced a mask-aware dynamic moving average aggregation strategy (MADA) that mitigates client drift. Additionally, \prism{} offers a hybrid mask/score aggregation method, allowing for a flexible and controllable trade-off between performance and efficiency. Our extensive experiments, including scenarios involving differential privacy and non-IID setups, demonstrated that \prism{} is robust in unstable environments. To the best of our knowledge, \prism{} is the first framework to consistently generate high-quality images with significantly reduced communication overhead in FL settings. 

\section*{Acknowledgments}
This work was supported by the National Research Foundation of Korea (NRF) grant funded by the Korea government (MSIT) (No.2022R1C1C1008496) and Institute of Information \& communications Technology Planning \& Evaluation (IITP) grant funded by the Korea government (MSIT): (No.2022-0-00959, No.RS-2022-II220959 (Part 2) Few-Shot Learning of Causal Inference in Vision and Language for Decision Making), (No.RS-2022-II220264, Comprehensive Video Understanding and Generation with Knowledge-based Deep Logic Neural Network), (No.RS-2020-II201336, Artificial Intelligence Graduate School Program (UNIST)), (No.RS-2021-II212068, Artificial Intelligence Innovation Hub).

\bibliography{iclr2025_conference}
\bibliographystyle{iclr2025_conference}

\appendix
\section{Appendix}
\appendix
\section{Privacy}
\label{appendix:dp}
Diffenrential privacy (DP) and R\'{e}nyi Differential Privacy (RDP)~\cit{dp, rdp} are the most popular definitions to analysis the privacy in FL environments. These help mitigate privacy concerns by limiting the contribution of individual data points. $(\epsilon, \delta)$-DP and $(\alpha, \epsilon)$-RDP basically calculates the distance of outcome for the algorithm of adjacent datasets.

\begin{definition}[($\epsilon, \delta$)-Differential Privacy]
A randomized mechanism $\mathcal{M} : \mathcal{X} \rightarrow \mathcal{R}$ is $(\epsilon, \delta)$-differential privacy, if for any two adjacent datasets $\mathcal{D}$, $\mathcal{D'}$ and for any measurable sets $\mathcal{S}$:
\begin{equation} \label{eq:dp}
    \textbf{Pr}[\mathcal{M}(D) \in \mathcal{S}] \leq e^{\epsilon} \textbf{Pr}[\mathcal{M}(D') \in \mathcal{S}] + \delta
\end{equation}
\end{definition}

\begin{definition}[($\alpha, \epsilon$) R\'{e}nyi Differential Privacy]
For two probability distributions P and Q, the R\'{e}nyi divergence of order $\alpha > 1$ defined as follows:
\begin{equation}
    R_{\alpha}(P||Q) \triangleq \frac{1}{\alpha-1} \log \mathbb{E}_{x\sim Q}(\frac{P(x)}{Q(x)})^{\alpha}
\end{equation}
\noindent then, a randomized mechanism $\mathcal{M} : \mathcal{X} \rightarrow \mathcal{R}$ is ($\alpha, \epsilon$) R\'{e}nyi differential privacy, if for any two adjacent datasets $\mathcal{D}$, $\mathcal{D'}$ and for any measurable sets $\mathcal{S}$:

\begin{equation}\label{eq:rdp}
    R_{\alpha}(\mathcal{M}(D)||\mathcal{M}(D')) \leq \epsilon
\end{equation}
\end{definition}

\begin{theorem}\cit{rdp} showed that if $\mathcal{M}$ is $(\alpha, \epsilon)$-RDP guarantee, is also $(\epsilon + \frac{\log 1 / \delta}{\alpha-1})$-DP.
    
\end{theorem}

In this section, we provide more detailed explanation of privacy preserving in PRISM and also present updated results when DP is applied. To satisfy the $(\epsilon, \delta)$-DP, our goal is privatize the probability vector $\theta \in [0, 1]^d$ by adding gaussian noise $\mathcal{N}(0,\sigma^2)$, where $\sigma^2=\frac{2ln(1.25/\delta)\Delta_2^2}{\epsilon^2}$ and $\Delta_2 = \max_{D,D'}||\mathcal{M}(D)-\mathcal{M}(D')||_2$. When the local training is end, each client has scores $s \in \mathbb{R}^d$ to choose which weight to prune. Recall that probability $\theta \in [0,1]^d$ can be obtained through sigmoid function. We inject gaussian noise and clip to $\tilde{\theta} \in [c, 1-c]^d$, where c is a small value $0<c<\frac{1}{2}$. In out setup, we fix it at 0.1. Now, we ensure $\tilde{\theta}$ is $(\epsilon, \delta)$-DP. For a fair comparison, we use $(\epsilon,\delta)=(9.8,10^{-5})$ to PRISM and our baselines in all of our experiments. In addition, we regulate the global round to ensure that the overall privacy budget does not exceed $\epsilon$. To track the overall privacy budget, we employ subsampled moments accountant~\cit{subsampled_moment_accountant}. We refer to the Opacus library which is the user-friendly pytorch framework for differential privacy~\cit{opacus}.

\cit{noisy_bernoulli, FedPM} have shown that performing post processing to already privatized vector $\tilde{\theta}$ such as Bernoulli sampling enjoys privacy amplification under some conditions. By doing so, the overall privacy budget becomes smaller $\epsilon_{amp} \leq \min \{\epsilon, d\gamma_{\alpha}(c)\}$, where $\gamma_{\alpha}(\cdot)$ is the binary symmetry R\'{e}nyi divergence as expressed below:

\begin{equation}\label{eq:binary_rdp}
    \gamma_{\alpha}(c)=\frac{1}{\alpha-1}\log (c^{\alpha}(1-c)^{\alpha}+(1-c)^{\alpha}c^{1-\alpha}),
\end{equation}
where $\alpha$ refers to the order of the divergence. Note that $d$ limits the privacy amplification when the model size becomes large. Since PRISM assumes that the model size is large enough due to SLT, we focus on communication efficiency rather than privacy amplification.

\section{Privacy amplified scenario}
\label{appendix:privacy}
In this section, we discuss about potential privacy benefit of \prism{} under some conditions. Typical FL setting, a malicious third party can estimate $W_t^k - W_h^g$ from the communicated gradients. By analyzing these gradients, an attacker can extract information about the local data. However, with \prism{}, only the binary masks $M_t^k$ and $M_t^g$ are exchanged, which hinders an attacker's efforts since they do not have access to the synchronized initial weight $W_{init}$ shared between the client and server.

\begin{table}[h]
\centering
\caption{Generator architecture used in our experiments.}
\begin{tabular}{@{}llcccc@{}}
\toprule
\textbf{Layer}         & \textbf{Type}          & \textbf{Input Channels} & \textbf{Output Channels} & \textbf{Kernel Size} & \\ 
\midrule
FirstConv & Conv & 128 & 512 & (4, 4)  \\
\midrule
\multirow{5}{*}{Resblock0} 
    & Conv1 & 512 & 256 & (3, 3) \\
    & Conv2 & 256 & 256 & (3, 3) \\
    & BatchNorm2d & 512 & 512 & \\
    & ReLU & - & - & - \\
    & Bypass Conv & 512 & 256 & (1, 1) \\
\midrule
\multirow{6}{*}{Resblock1}
    & Upsample & - & - & -  \\
    & Conv1 & 512 & 256 & (3, 3) \\
    & Bypass Conv & 512 & 256 & (1, 1) \\
    & Upsample & - & - & -  \\
    & Conv2 & 256 & 128 & (3, 3)  \\
    & Conv3 & 128 & 128 & (3, 3)  \\
\midrule
\multirow{4}{*}{Resblock2}
    & Conv1 & 128 & 64 & (3, 3) \\
    & Conv2 & 64 & 64 & (3, 3)  \\
    & BatchNorm2d & 64 & 64 & -  \\
    & ReLU  & -   & -  & -   \\
\midrule
\multirow{2}{*}{LastConv}
    & Conv & 64  & 1  & (3, 3)  \\
    & Tanh   & -   & -   & -  \\ 
\bottomrule
\end{tabular}
\label{tab:architecture}
\end{table}

\section{Training Details}
\label{appendix:training_details}

In this section, we provide the detailed description of our implementations and experimental settings. In \Tref{tab:architecture}, we provide the model architectures used in our experiments. We use ResNet-based generator and set the local epoch to 100 and learning rate to 0.1. In addition, we do not employ training schedulers or learning rate decay. Our code is based on~\cit{gfmn, ep_mmd}. They employ the ImageNet-pretrained VGG19 network for feature matching by minimizing the Eq.~\ref{eq:mmd2}. However, calculating the first and second moments require the large batch size to obtain the accurate statistics. To address this issue,~\cit{gfmn} introduces Adam moving average (AMA). With a rate $\lambda$, the update of AMA $m$ is expressed as follows:
\begin{equation}
    m \leftarrow m - \lambda \text{ADAM}(m-\Delta),
\label{eq:AMA_1}
\end{equation}
where $\text{ADAM}$ denotes Adam optimizer~\cit{adam} and $\Delta$ is the discrepancy of the means of the extracted features. Note that $\text{ADAM}(m-\Delta)$ can be interpreted as gradient descent by minimizing the L2 loss:
\begin{equation}
    \underset{m}{\min}\frac{1}{2} \left\|m-\Delta \right\|^2.
\label{eq:AMA_2}
\end{equation}
This means the difference of statistics $(m-\Delta)$ is passed through a single MLP layer and updated using the Adam optimizer to the direction of minimizing Eq.~\ref{eq:AMA_2}. By utilizing AMA, Eq.~\ref{eq:mmd2} is formulated as $\mathcal{L}_{MMD}^k=\left\| \mathbb{E}_{x\sim \mathcal{D}^k}[\psi(x)] - \mathbb{E}_{y\sim \mathcal{D}_{fake}^k}[\psi(y)]\right\|^2 + \left\| \Cov(\psi(\mathcal{D}^k)) - \Cov(\psi(\mathcal{D}_{fake}^k))\right\|^2,$
Algorithm~\ref{alg:prism},~\ref{alg:prism-alpha} provides the pseudocode for \prismdema{} and \prism{}$^{\ast}$ correspondingly. AMA is omitted to simply express the flow of our framework. See our \href{https://github.com/tjrudrnr2/PRISM_ICLR25}{code} for pytorch implementation. We train the local generator for 100 local iterations with learning rate of 0.1. For the AMA layer, learning rate is set to 0.005. In addition, we use the Adam optimizer with $\beta_1=0.5, \beta_2=0.999$ to update the scores of the generators. After all clients complete their training, communication round is initiated. We set the global epoch to 150 for the MNIST dataset and 350 for the CelebA and CIFAR10 datasets. As we do not adjust the parameters, note that there is room for performance improvements through hyperparameter tuning.

\renewcommand{\algorithmicrequire}{\textbf{Input:}}
\renewcommand{\algorithmicensure}{\textbf{Parameter:}}
\begin{algorithm}[h]
\caption{\prismdema{}}
\label{alg:prism}
\begin{algorithmic}[t]
\Ensure{learning rate $\eta$, communication rounds $\textit{T}$, local iterations $\textit{I}$}
\Require{local datasets $\cup_{k=1}^K\mathcal{D}^k$}, ImageNet pretrained VGGNet $\psi$, random noise $z$
\Statex
\State \textbf{Server execute:}
\State Initialize a random weight $W_{init}$ and score vector $s$, then broadcasts to all clients.
\For{\textit{round t = $1,...,T$}}
    \State \textbf{Client side:}
    \For{\textit{each client $k \in \left[1,K\right]$}}
        \State $s_t^k = s_t$ \Comment{Download score vector}
        \For{\textit{local iteration $i =1,,,L$}}
            \State $\theta_t^k \leftarrow \text{Sigmoid}(s_t^k)$
            \State $M_t^k \sim \text{Bern}(\theta_t^k)$
            \State $W_t^k \leftarrow W_{init} \odot M_t^k$
            \State $\mathcal{D}_{fake}^{k}$ $\leftarrow W_t^k(z)$ \Comment{Generate fake images}
            \State Extract real and fake features $\psi({\mathcal{D}^k}), \psi({\mathcal{D}_{fake}^k})$
            \State $s_t^k \leftarrow s_t^k -\eta \nabla \mathcal{L}_{MMD}^k(\psi({\mathcal{D}^k}), \psi({\mathcal{D}_{fake}^k}))$ \Comment{Update local score vector}
        \EndFor
        \State $\bar{\theta_t^k} \leftarrow \text{Sigmoid}(s_t^k)$
        \State $\tilde{\theta_t^k} = \leftarrow \bar{\theta_t^k} + \mathcal{N}(0, I\sigma^2)$ 
        \State \text{Clip to [c, 1-c]}
        \State $M_t^k \sim \text{Bern}(\theta_t^k))$
        \State Upload binary mask $M_t^k$ to the server.
    \EndFor
    \State \textbf{Server side:}
    \State $\hat{\theta}_{t+1} \leftarrow \sum_{k=1}^K M_t^k$ \Comment{Aggregate the received binary masks}
    \State $s_{t+1} \leftarrow \text{Sigmoid}^{-1}(\hat{\theta}_{t+1})$
    \State $\lambda \leftarrow hd(M_{t}, Bern(\hat{\theta}_{t+1}))$ \Comment{Compute the hamming distance}
    \State $s_{t+1} \leftarrow (1-\lambda)s_t + \lambda s_{t+1}$
\EndFor
\State Sample the supermask $M^* \sim \text{Bern}(\theta_T)$
\State Obtain the final model $W^*\leftarrow W_{init}\odot M^*$
\end{algorithmic}
\end{algorithm}

\renewcommand{\algorithmicrequire}{\textbf{Input:}}
\renewcommand{\algorithmicensure}{\textbf{Output:}}
\begin{algorithm}[h]
\caption{\prism{}$^{\ast}$}
\label{alg:prism-alpha}
\begin{algorithmic}[t]
\Require{ratio of score layer $\alpha$}
\Ensure{probability $\theta_t^k(\100-\alpha)$ and binary mask $M_t^k(\alpha)$}
\Statex
\State \textbf{Client side:}
\For{\textit{layer l = $1,...,L$}}
    \If{IsScoreLayer(l,$\alpha$,L)}
        \State Return probability $\theta_t^k(l)$
    \Else
        \State Return binary mask $M_t^k(l)$
    \EndIf
\EndFor
\end{algorithmic}
\end{algorithm}

\section{Dataset partitioning strategies across clients}
\label{appendix:NonIID_detail}
Here, we elaborate on the dataset partitioning strategies used to simulate the IID and non-IID setup. For non-IID setup, we used two strategies for partitioning datasets: 1) Shards-partitioned and 2) Dirichlet-partitioned. 

\noindent\textbf{IID setup.} 
In the case of IID scenario, we assign equal-sized local datasets by uniformly sampling from the entire training set. This allows for a balanced distribution of data across the clients. 

\begin{table}[h]
\caption{\textbf{Quantitative comparison in Dirichlet non-IID scenario with $(\epsilon, \delta)=(9.8, 10^{-5})$.} We compare FID, P\&R, D\&C. $\dagger$ indicates that \prismdema{} is removed.
}
\vspace{-2mm}
\begin{center}
\resizebox{0.8\textwidth}{!}{
\begin{tabular}{ccccc}
\toprule
\multirow{1}{*}{\textbf{Method}} & \multirow{1}{*}{\textbf{Metric}} & \multirow{1}{*}{\textbf{MNIST}} & \multirow{1}{*}{\textbf{FMNIST}} & \multirow{1}{*}{\textbf{CelebA}} \\
\midrule
\multirow{3}{*}{\begin{tabular}[c]{@{}c@{}}
    GS-WGAN \end{tabular}}
                        & \textbf{FID $\downarrow$} & 128.4401 & 134.4054 &  108.8792  \\
                        & \textbf{P\&R $\uparrow$} & 0.0851 / 0.0633 & 0.3927 / 0.0001 & \textbf{0.8891} / 0.0154   \\
                        & \textbf{D\&C $\uparrow$} & 0.0196 / 0.0071 & \textbf{0.1252} / 0.0348 & 0.2165 / 0.052  \\
\midrule
\multirow{3}{*}{\begin{tabular}[c]{@{}c@{}}
    DP-FedAvgGAN  \end{tabular}}
                        & \textbf{FID $\downarrow$} & 175.3729 & 209.1346 & 323.3559  \\
                        & \textbf{P\&R $\uparrow$} & 0.0408 / \textbf{0.1982} & 0.0814 / \textbf{0.0421} & 0.1357 / 0.0  \\
                        & \textbf{D\&C $\uparrow$} & 0.0102 / 0.0048 & 0.0214 / 0.0123 & 0.0271 / 0.0001  \\
\midrule
\multirow{3}{*}{\begin{tabular}[c]{@{}c@{}} \prism{}$^{\dagger}$  \end{tabular}}
    & \textbf{FID $\downarrow$} & 74.739 & 137.7545 & 57.4051  \\
    & \textbf{P\&R $\uparrow$} & 0.2209 / 0.1156 & 0.3041 / 0.0274 & 0.482 / \textbf{0.1058}  \\
    & \textbf{D\&C $\uparrow$} & 0.0619 / 0.0415 & 0.1086 / 0.0312 & 0.2317 / 0.1934 \\
\midrule
\multirow{3}{*}{\begin{tabular}[c]{@{}c@{}} \prism{}  \end{tabular}}
    & \textbf{FID $\downarrow$} & \textbf{70.9146} & \textbf{137.7359} & \textbf{38.0283} \\
    & \textbf{P\&R $\uparrow$} & \textbf{0.4119} / 0.1563 & \textbf{0.3394} / 0.0378 & 0.6305 / 0.0746  \\
    & \textbf{D\&C $\uparrow$} & \textbf{0.1513} / \textbf{0.0759} & 0.1139 / \textbf{0.0394} & \textbf{0.4493} / \textbf{0.3144}  \\
\bottomrule
\end{tabular}
}
\label{tab:noniid-dir_dp}
\end{center}
\end{table}

\begin{table}[h]
\caption{\textbf{Quantitative comparison in Dirichlet non-IID scenario without DP.} We compare FID, P\&R, D\&C. $\dagger$ indicates that \prismdema{} is removed. We set $\alpha=80$ for \prism{}$^{\ast}$.}
\vspace{-2mm}
\begin{center}
\resizebox{\textwidth}{!}{
\begin{tabular}{cccccc}
\toprule
\multirow{1}{*}{\textbf{Method}} & \multirow{1}{*}{\textbf{Metric}} & \multirow{1}{*}{\textbf{MNIST}} & \multirow{1}{*}{\textbf{FMNIST}} & \multirow{1}{*}{\textbf{CelebA}} & \multirow{1}{*}{\textbf{CIFAR10}} \\
\midrule
\multirow{3}{*}{\begin{tabular}[c]{@{}c@{}}
  MD-GAN  \end{tabular}} 
                        & \textbf{FID $\downarrow$} & 106.3468 & 86.0443 & 30.7787 & \textbf{54.38} \\
                        & \textbf{P\&R $\uparrow$} & \textbf{0.7109} / 0.2431 & 0.5586 / 0.0359 & 0.7612 / \textbf{0.6425} & \textbf{0.8941} / 0.1903 \\
                        & \textbf{D\&C $\uparrow$} & \textbf{0.6884} / \textbf{0.4895} & 0.2137 / 0.0917 & 0.7238 / 0.4267 & \textbf{1.4087} / \textbf{0.355} \\
\midrule
\multirow{3}{*}{\begin{tabular}[c]{@{}c@{}}
  Multi-FLGAN  \end{tabular}} 
                        & \textbf{FID $\downarrow$} & \multicolumn{4}{c}{\multirow{3}{*}{diverge}} \\ 
                        & \textbf{P\&R $\uparrow$}  \\  
                        & \textbf{D\&C $\uparrow$}  \\
\midrule
\multirow{3}{*}{\begin{tabular}[c]{@{}c@{}} \prism{}$^{\dagger}$ \end{tabular}}
    & \textbf{FID $\downarrow$} & 43.3057 & 76.1413 & 21.0935 & 89.0257  \\
    & \textbf{P\&R $\uparrow$} & 0.6186 / 0.216 & \textbf{0.6093} / 0.023 & \textbf{0.797} / 0.2229 & 0.6916 / 0.0854  \\
    & \textbf{D\&C $\uparrow$} & 0.3219 / 0.1799 & \textbf{0.3443} / 0.1163 & 0.991 / 0.6533 & 0.5885 / 0.2157  \\
\midrule
\multirow{3}{*}{\prism{}  }
                        & \textbf{FID $\downarrow$} & \textbf{31.6191} & 73.8701 & 20.0883 & 74.1502  \\
                        & \textbf{P\&R $\uparrow$} & 0.5871 / 0.36 & 0.5239 / 0.0378 & 0.782 / 0.2024 & 0.572 / 0.1601 \\
                        & \textbf{D\&C $\uparrow$} & 0.2828 / 0.2328 & 0.2919 / 0.1054 & \textbf{1.1148} / 0.6893 & 0.3838 / 0.2535 \\
\midrule
\multirow{3}{*}{\prism{}$^{\ast}$ }
    & \textbf{FID $\downarrow$} & 38.8984 & \textbf{72.8571} & \textbf{13.6906} & 57.9337  \\
    & \textbf{P\&R $\uparrow$} & 0.5671 / \textbf{0.5423} & 0.4926 / \textbf{0.045} & 0.784 / 0.3266 & 0.6395 / \textbf{0.2645}  \\
    & \textbf{D\&C $\uparrow$} & 0.27 / 0.2251 & 0.2628 / \textbf{0.1177} & 0.9627 / \textbf{0.7352} & 0.567 / 0.327  \\
\bottomrule
\end{tabular}
}
\label{tab:noniid-dir}
\end{center}
\vspace{-4mm}
\end{table}

\noindent\textbf{Non-IID setup (Shards-partitioned).} 
In~\Sref{subsec:noniid} and~\Sref{subsec:noniid w/o dp}, we partition the MNIST, FMNIST, and CIFAR10 datasets into 40 segments based on the sorted class labels and randomly assign four segments to each client. 
For the CelebA dataset, which contains multiple attributes per image, defining a clear non-IID distribution for splitting is inherently ambiguous. In this work, we divide the dataset into tow partitions based on a pivotal attribute (gender, in our case). The total number of partitions corresponds to the number of clients, with each client being assigned either the positive or negative subset of images for the pivotal attribute. However, the remaining 39 attributes are still shared among clients, resulting in relatively weak heterogeneity. This explains why the performance drop on CelebA is not as significant compared to the IID case. For a more detailed discussion, please refer to~\Aref{appendix:NonIID_detail}.

\noindent\textbf{Non-IID setup (Dirichlet-partitioned).} 
We further explore a more realistic and challenging non-IID scenario, where datasets are partitioned using Dirichlet distribution. Specifically, we set Dirichlet parameter $\alpha=0.005$ to create a more label-skewed distribution. For CelebA, we assign data to clients such that each client possesses the pivotal attribute in different proportions using Dirichlet distribution. For example, client 1 has 60\% male and 40\% female, while client 2 has 20\% male and 80\% female. To the best of our knowledge, methods for splitting multi-label datasets into extreme non-IID scenarios have not been extensively explored. \Tref{tab:noniid-dir_dp} and~\Tref{tab:noniid-dir} present a quantitative comparison of baselines and our methods under both with and without DP. \prism{} demonstrates robust performance even under highly heterogeneity distributions. Note that Multi-FLGAN fails to successfully train generative model under highly heterogeneity.

\section{Efficient Storage and Quantization Strategy in~\prism{}}
\label{appendix:quantization}
In this section, we provide further detail description regarding storage efficiency of~\prism{}. Aforementioned in~\Sref{subsec:PRISM}, the final model can be expressed as $W^*=W_{init}\odot M^*$. Since $M_{fianl}$ is lightweight binary masks, we need to focus on the storage of $W_{init}$. Note that $W_{init}$ is initialized as signed constant using Kaiming Normal distribution. In other words, the initialized parameter of layer $l$ becomes $\{+\sqrt{2/n_{l-1}}, -\sqrt{2/n_{l-1}}, \cdots, +\sqrt{2/n_{l-1}} \}\in \mathbb{R}^d$. This design makes storing the scaling factor $v=\sqrt{2/n_{l-1}}$ be enough to represent each parameter, represented as $W_{init}=v \odot M_{sign}, \text{where } M_{sign}$ is defined as $\{+1, -1, \cdots, +1 \}\in \mathbb{R}^d$. Our memory-efficient approach is closely related to ternary quantization~\cit{ternary_quantization}, which typically compresses neural network weights into $\{-1, 0, 1\}$ through thresholding and projection. However,~\prism{} distinguishes itself by directly initialized frozen weights as signed constants using the Kaiming Normal distribution. Since storing a signed array is negligible in terms of memory overhead, this design makes the final model exceptionally lightweight without extra pruning or quantization.

\section{Analysis of hybrid aggregation}
\label{appendix:hybrid}

\begin{figure}[t]
  \centering
  \includegraphics[width=\textwidth]{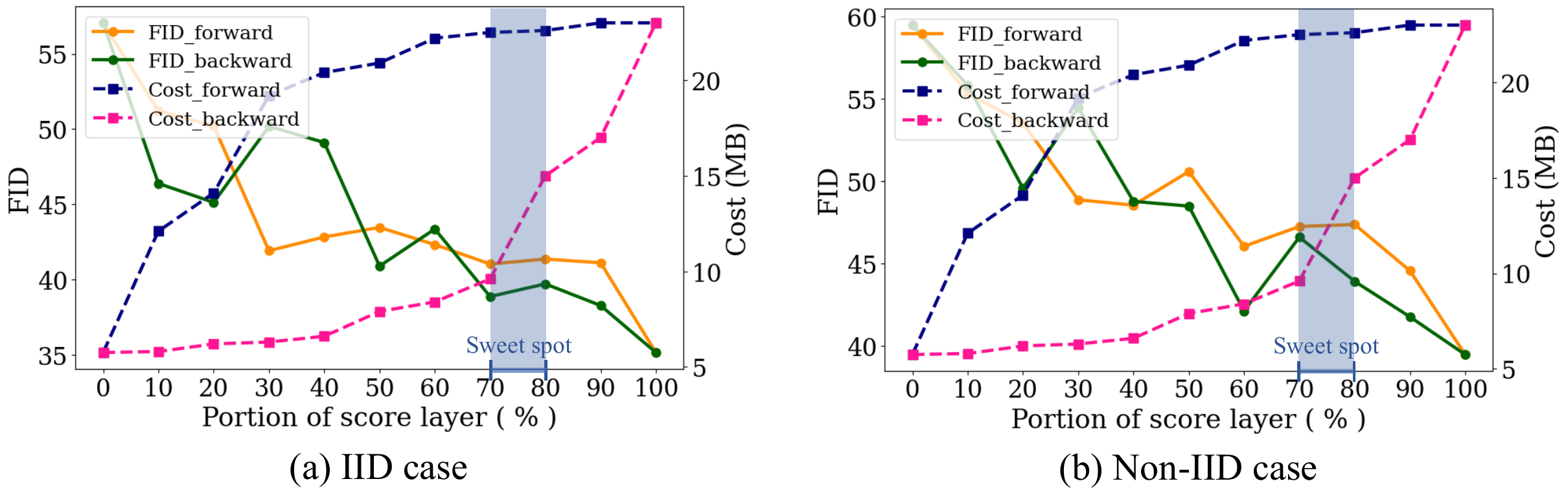}
  \caption{\textbf{Analysis of \prism{}$^{\ast}$~using CIFAR10 dataset.} The \textit{backward path} selects $\alpha\%$ of \textit{score layers} from deeper layers, closer to the output, while the \textit{forward path} chooses from the opposite end, nearer to the input. Solid-line demonstrates FID following each direction while dash-line shows communication cost (MB) of each path.}
  \label{fig:forward_backward}
\end{figure}

\begin{figure}[t]
  \centering
  \includegraphics[width=\textwidth]{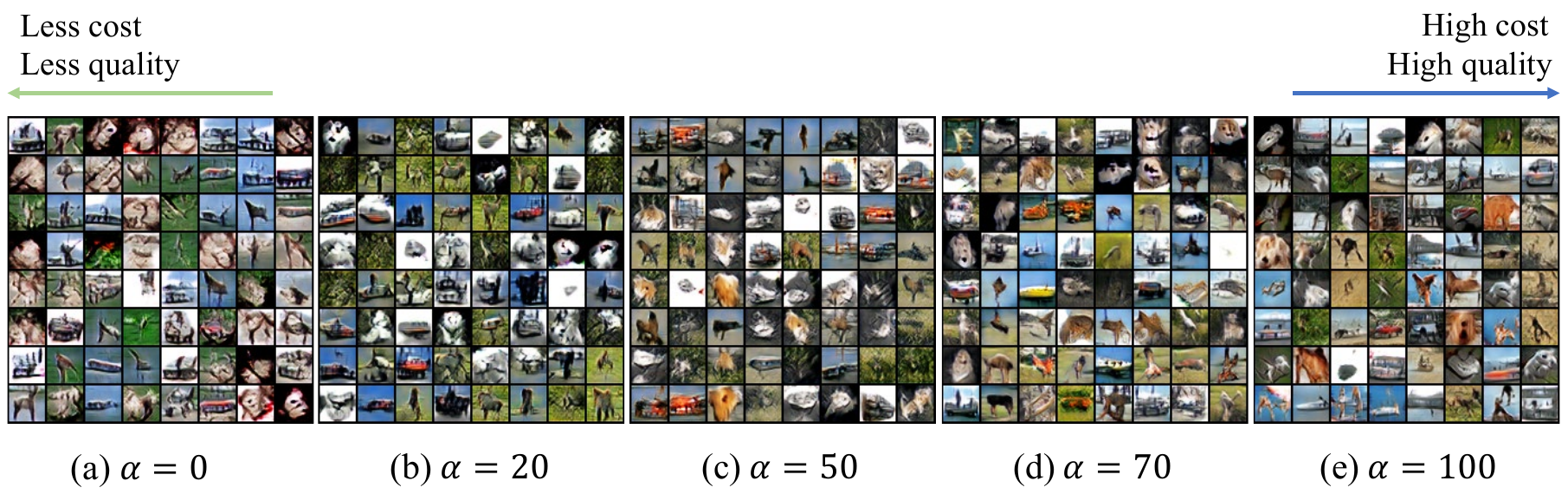}
  \caption{\textbf{The effect of adjusting $\alpha$ of \prism{}$^{\ast}$.} Qualitative comparison according to $\alpha$.
  Note that $\alpha=0$ is identical to \prism{}.} 
  \label{fig:portion_ablation}
     \vspace{-3mm}
\end{figure}

In this section, we further analyze \prism{}$^{\ast}$, which leverages both binary mask and score communications. To explore the trade-off between communication cost and generative capability, we consider two strategies: the \textit{backward path} and \textit{forward path}. The \textit{backward path} progressively increases the number of \textit{score layers} from deeper layers to earlier layers. Conversely, in the \textit{forward path}, we select \textit{score layers} from earlier layers to deeper layers. \Fref{fig:forward_backward} visually demonstrates the trade-off between communication cost and FID of both strategies. The FID gradually improves as we increase $\alpha$ values in both cases. Note that the additional communication cost of the \textit{backward path} tends to increase more smoothly. Based on these observations, we adopt the \textit{backward path} for \Tref{tab:noniid} and \Fref{fig:noniid}. \Fref{fig:portion_ablation} provides a comprehensive comparison across a wide range of $\alpha$ values, showing that \prism{}$^{\ast}$ consistently produces high quality images. Note that all experiments are conducted in privacy-free scenario.

\section{Communication cost during downlink process}
\label{appendix:discussion_cost}
In a typical FL setup, the server possesses powerful computational capabilities, whereas the clients do not. As a result, several prior studies primarily focus on addressing the limited bandwidth of clients~\cit{uplink1, uplink2, uplink3}. Accordingly, we have concentrated on the reducing communication cost during the uplink process, transmitting a float-type score during downlink to mitigate the information loss. Last but not least, binary mask communication can also be employed in the downlink to further enhance communication efficiency, despite the occurrence of information loss during the conversion of the global mask into local scores.

\begin{table}[h]
\caption{\textbf{Quantitative results on cross-device environment.} We set the number of clients to 50 and participant ratio to 20\%.}
\begin{center}
\resizebox{\columnwidth}{!}{
\begin{tabular}{ccccc||c}
\toprule
\textbf{Case} & \textbf{Metric} & \textbf{MD-GAN} & \textbf{DP-FedAvgGAN} & \textbf{GS-WGAN} & \textbf{PRISM} \\ 
\midrule
\multirow{3}{*}{\begin{tabular}[c]{@{}c@{}}
  Non-IID w/ DP \end{tabular}} 
                         & \textbf{FID $\downarrow$} & N/A & 118.3975 & 98.6553 & \textbf{34.7157} \\ 
                        & \textbf{P\&R $\uparrow$} & N/A & 0.1095 / 0.3723 & \textbf{0.8477} / 0.0359 & 0.4344 / \textbf{0.3401} \\ 
                        & \textbf{D\&C $\uparrow$} & N/A & 0.0301 / 0.0289 & \textbf{0.2621} / 0.0105 & 0.1692 / \textbf{0.1476} \\ 
\midrule
\multirow{3}{*}{\begin{tabular}[c]{@{}c@{}}
  Non-IID w/o DP \end{tabular}} 
                        & \textbf{FID $\downarrow$} & 15.4119 & N/A & N/A & \textbf{14.3168} \\ 
                        & \textbf{P\&R $\uparrow$} & 0.7305 / 0.359 & N/A & N/A / N/A & \textbf{0.7533} / \textbf{0.4757} \\ 
                        & \textbf{D\&C $\uparrow$} & 0.5266 / 0.3803 & N/A & N/A & \textbf{0.5804} / \textbf{0.5293} \\ 
\bottomrule
\end{tabular}
}
\label{tab:crossdevice_dp}
\end{center}
\end{table}

\section{Cross-device environments}
\label{appendix:crossdevice}
\prism{} is designed to provide efficient communication cost, making it ideal for cross-device environments with numerous clients and limited bandwidth. Typically, cross-device setting involves a large number of clients and an unreliable network, such as those encountered in iPhone and wearable devices. This complexity makes it difficult to ensure that all clients in every communication round. To simulate this cross-device environment, we set the number of clients to 50 and the participant ratio as 20\%. \Tref{tab:crossdevice_dp} shows the quantitative results using MNIST dataset, both with and without DP. \prism{} demonstrates a clear performance advantage over other baselines and is well-suited for the cross-device environment.

\section{Analysis of convergence of $\lambda$ in MADA}
\label{appendix:Analysis_MADA}
In this section, we provide a detailed discussion on how the reduction in $\lambda$, which measures the similarity between two consecutive global masks, helps mitigate client drift. 

One possible explanation is that MADA implicitly modulates the learning rate for global updates on the server-side without requiring an explicit learning rate scheduler. This behavior reduces the need for client-side learning rate scheduling to regularize local objectives. To illustrate this, consider a local mask update represented as $w_k^t=w_k^{t-1}-\eta \cdot \Delta_k^t$, where $\eta$ is constant learning rate, and $\Delta_k^t$ is local model update from client $k$. In standard FedAvg, the global update can be expressed as $w_{\text{FedAvg}}^t = w_k^{t-1} -\eta \cdot \Sigma_k \Delta_k^t$, which aggregates all client updates equally. In contrast, MADA introduces a mask-aware aggregation mechanism, and the global model is formulated as $w_{\text{MAGA}}^t=(1-\lambda)w_k^{t-1}+\lambda \cdot w_k^t$, where $w_k^t=w_k^{t-1}-\eta \cdot \Delta_k^t$. Substituting the local mask update into this equation, $w_{\text{MAGA}}^t=(1-\lambda)w_k^{t-1}+\lambda \cdot (w_k^{t-1}-\eta \cdot \Sigma_k \Delta_k^t)=w_k^{t-1}-\eta \cdot \lambda \Sigma_k \Delta_k^t$, indicating that $\lambda$ effectively scales the global updates step size. When $\lambda$ is large, the global model closely follows the client updates, while a small $\lambda$ reduces the influence of client updates, acting as a dampening factor.

The intuition behind this behavior is rooted in the nature or early-stage training. During the initial rounds, the model is far from convergence, resulting in large gradient norms and substantial parameter changes. Consequently, consecutive global masks exhibit significant differences, leading to a large $\lambda$. As training progresses and the model stabilizes, mask updates become more similar, decreasing $\lambda$. This gradual reduction in $\lambda$ prevents excessive global updates, stabilizing convergence and mitigating client drift. We elaborate on the behavior of client drift. Even when client-drift is biased toward a specific dominant client, the update difference for the dominant client naturally decreases from the perspective of global support, whereas the difference from the local solutions of other clients increases. This pattern is reflected in the aggregated model, and the difference between consecutive global models can partially capture this behavior (e.g., indicated by a large $\lambda$). As training progresses and the model approaches the global optimum across local objectives, the differences between consecutive global models gradually decrease, leading to a reduction in $\lambda$. Conversely, in the absence of client drift, such as in IID scenarios, the model learns fairly across all clients and naturally converges without notable fluctuations in global model differences.

In summary, MADA updates the global model in a way that avoids local divergence, gradually reducing global model differences even without learning rate decay. As shown in~\Fref{fig:local-divergence}, $\lambda$ decreases sharply in the early stages of training due to significant local divergence but gradually converges over time. This result suggests that our understanding aligns with the behavior of $\lambda$.

\begin{table}[h]
\caption{Comparison of FLOPS between~\prism{} and other baselines.}
\begin{center}
\resizebox{0.55\columnwidth}{!}{
\begin{tabular}{cccc}
\toprule
 & \textbf{DP-FedAvgGAN} & \textbf{GS-WGAN} & \textbf{\prism{}} \\ 
\midrule
  FLOPS & 0.002 B & 1.94 B & 0.34 B \\
\bottomrule
\end{tabular}
}
\label{tab:flops}
\end{center}
\end{table}
\section{Computation cost analysis}
\label{appendix:FLOPS}

One might say that SLT process within PRISM increases computational complexity, particularly in large networks. As the number of model parameters grows, the required score parameters also scale linearly. However, in SLT, weight parameters are not optimized via gradient descent. The only additional computations stem from the sigmoid and Bernoulli processes introduced during the binary masking procedure. While these operations introduce some additional computational overhead, they remain negligible in the context of GPU-accelerated gradient descent. To demonstrate that the additional computational cost is not substantial, we compare floating-point-operations per second (FLOPS) in~\Tref{tab:flops}. GS-WGAN exhibits increased FLOPS due to spectral normalization, which is applied to stabilize GAN training. In contrast, \prism{}, despite incorporating the SLT process, maintains a sufficiently low FLOPS, ensuring its computational efficiency.

\begin{figure}[h]
  \centering
  \includegraphics[width=0.3\textwidth]{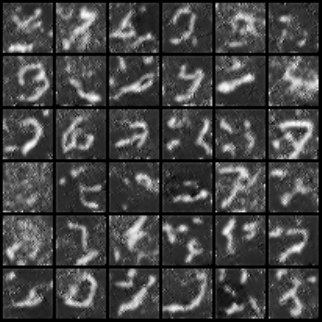}
  \caption{Generated images of \prism{} where ddpm is used instead of MMD loss.}
  \label{fig:prism_ddpm}
\end{figure}
\section{Scalability of \prism{}}
Due to the challenges associated with training generative models in FL setups, research on federated generative models has primarily focused on simple architectures (\textit{e.g.,} GANs) and datasets, such as MNIST and FMNIST level generation. However, it is important to note that the field of federated generative models remains far behind the current trend of using more complex and powerful models, such as diffusion models, and large-scale datasets. This gap arises from the inherent difficulties in adapting these advanced models to FL setups. Existing works have struggled to achieve stable performance in challenging setups, such as non-IID and privacy-preserving cases, even with MNIST-level datasets. Even for centralized differential privacy generative models~\cit{centralized_dp_gen_1, centralized_dp_gen_2},training generative models is highly challenging and often unstable. Due to these obstacles, prior studies have primarily focused on simpler datasets, such as MNIST, and have been largely restricted to GANs.

\subsection{Extend to Diffusion model}
\label{appendix:prism_ddpm}
We further investigate the applicability of~\prism{} to larger and mode powerful generative models, such as diffusion models. Specifically, we retain the core concept of identifying SLT but replaced the MMD loss with DDPM loss~\cit{ddpm} to guide score updates. Notably, all other configurations remain unchanged. We conduct experiments on the MNIST dataset under a non-IID, privacy-free scenario. \Fref{fig:prism_ddpm} presents the generated images on the server-side, which exhibit slight noisiness. We anticipate that by improving the approach for identifying SLT within diffusion models and further curating the experimental setup, we can achieve more promising results. On the other hand, directly applying the MMD loss-based approach used in PRISM, which requires generating a set of samples for each update, poses practical challenges due to the well-known limitations of DDPM, including slow sampling and convergence rates. These limitations lead to significant computational costs, making the approach practically infeasible. Since our primary focus is lightweight communication cost and stable performance, we leave this future work, efficiently incorporating the powerful and diffusion model in federated settings.

\subsection{Evaluation on large-scale datasets}
\label{appendix:largescale}
\begin{figure}[h!]
    \centering
    \begin{minipage}{0.45\textwidth}
        \centering
        \includegraphics[width=\textwidth]{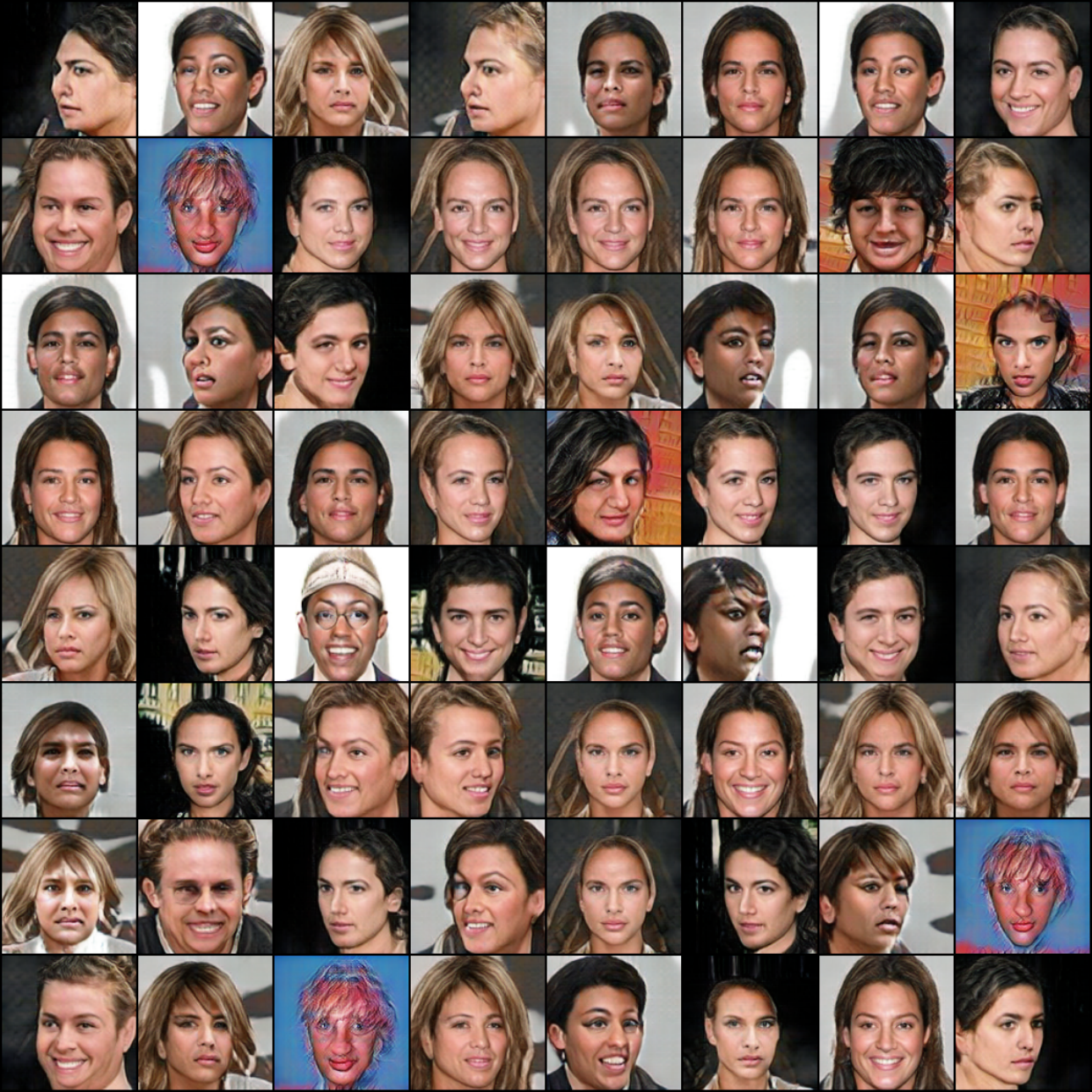}
    \end{minipage}
    \hfill
    \begin{minipage}{0.45\textwidth}
        \centering
        \includegraphics[width=\textwidth]{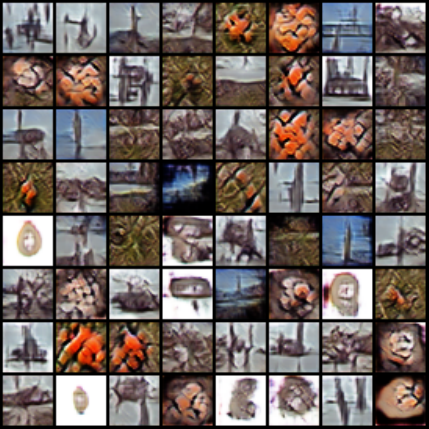}
    \end{minipage}
    \caption{Generated images of CelebA 128x128 dataset and CIFAR100 dataset under non-IID and privacy-free scenario.}
    \label{fig:largescale_dataset}
\end{figure}

\begin{table}[h!]
\caption{Performance of \prism{} using CelebA 128x128 and CIAFR100 datasets under non-IID and privacy-free scenario.}
\begin{center}
\resizebox{0.6\columnwidth}{!}{
\begin{tabular}{cccc}
\toprule
\textbf{Dataset} & \textbf{FID} & \textbf{P\& R} & \textbf{D\& C} \\ 
\midrule
CelebA 128 & 40.2927 & 0.7738 / 0.006 & 1.0027 / 0.348 \\ 
\midrule
CIFAR100 & 74.1609 & 0.6655 / 0.0719 & 0.602 / 0.3121 \\ 
\bottomrule
\end{tabular}
}
\label{tab:largescale_dataset}
\end{center}
\end{table}
Note that the field of federated generative models have struggled to achieve stable performance in challenging setups, such as non-IID and privacy-preserving cases, even with MNIST-level datasets. However, even under these conditions,~\prism{} is the first to consistently demonstrate stable performance under these conditions. Despite the poor conditions of federated generative models, it is important to address whether~\prism{} can be extended to large-scale datasets under limited resources. We provide quantitative and qualitative results on large-scale datasets in~\Tref{tab:largescale_dataset} and ~\Fref{fig:largescale_dataset}, such as CelebA 128x128 and CIFAR100 datasets, under non-IID and privacy-free scenario. Both experiments are conducted under exactly same settings as in~\Sref{sec:experiments}. As a result,~\prism{} demonstrates acceptable performance on high-resolution images; however, it reveals certain limitations when applied to large-scale datasets with diverse classes, such as CIFAR100. 

\section{Selection of other distance metric}
\label{appendix:cos}

\begin{table}[h!]
\caption{FID comparison where distance metric is cosine similarity for MADA.}
\vspace{-2mm}
\begin{center}
\resizebox{0.8\columnwidth}{!}{
{\scriptsize
\begin{tabular}{ccccc}
\toprule
\multirow{1}{*}{\textbf{Distance}} & \multirow{1}{*}{\textbf{Case}} & \multirow{1}{*}{\textbf{MNIST}} & \multirow{1}{*}{\textbf{FMNIST}} & \multirow{1}{*}{\textbf{CelebA}}  \\
\midrule
\prism{} w/ hd & \multirow{2}{*}{IID, DP} & \textbf{27.3017} & 46.1652 & \textbf{48.9983} \\
\prism{} w/ cos &    & 27.7895 & \textbf{44.4084} & 49.0747 \\
\midrule
\prism{} w/ hd & \multirow{2}{*}{Non-IID, DP} & \textbf{34.2038} & \textbf{67.1648} & \textbf{39.799} \\
\prism{} w/ cos &    & 34.9577 & 69.2994 & 51.1734 \\
\bottomrule
\end{tabular}
}
}
\label{tab:cos}
\end{center}
\end{table}

\Tref{tab:cos} provide the FID performance of \prism{} when using cosine similarity instead of hamming distance for \prismdema{}. \prism{} with cosine similarity achieves competitive performance compared to the scheme using Hamming distance in most cases. Although the performance degrades for CelebA in the non-IID scenario, it still outperforms all baseline methods, further demonstrating the robustness of the approach.

\section{The effect of Masking mechanism}
\label{appendix:selection_masking}

\begin{table}[h]
\caption{Effect of mask selection on MNIST.}
\begin{center}
\resizebox{0.8\columnwidth}{!}{
\begin{tabular}{ccccc||c}
\toprule
\textbf{Dataset} & \textbf{Metric} & \textbf{Random} & \textbf{Top-k\%} & \textbf{Bernoulli (ours)} & \textbf{Weight} \\ 
\midrule
\multirow{3}{*}{\begin{tabular}[c]{@{}c@{}}
  MNIST \end{tabular}} 
                        & \textbf{FID $\downarrow$} & 391.8257 & 20.3616 & \textbf{12.7373} & 5.9895 \\ 
                        & \textbf{P\&R $\uparrow$} & 0.0 / 0.0006 & 0.5232 / 0.3782 & \textbf{0.7323} / \textbf{0.5904} & 0.6783 / 0.8414 \\ 
                        & \textbf{D\&C $\uparrow$} & 0.0 / 0.2511 & 0.2854 / 0.0 & \textbf{0.5556} / \textbf{0.5313} & 0.446 / 0.678 \\ 
\bottomrule
\end{tabular}
}
\label{tab:mask_selection}
\end{center}
\end{table}
In \Tref{tab:mask_selection}, we compare the effect of the mask extraction algorithm on generative model training. For this, we relied on prior studies on the existence of SLT and the convergence of MMD with a characteristic kernel [46, 14]. Our results empirically show SLT’s convergence under mask averaging generative FL setting. We also explore the Random and Top-k\% algorithms. The Bernoulli method employed in PRISM outperformed top-k\%, while Random and Weight represent the lower and upper bounds of performance achievable by SLT, respectively.

\section{Evaluation on the Centralized Setting}
\label{appendix:centralized_setting}
\begin{table}[h]
\caption{Comparison of \prism{} with centralized setting.}
\begin{center}
\resizebox{0.8\columnwidth}{!}{
\begin{tabular}{ccccc}
\toprule
\multirow{1}{*}{\textbf{Method}} & \multirow{1}{*}{\textbf{Metric}} & \multirow{1}{*}{\textbf{MNIST}} & \multirow{1}{*}{\textbf{FMNIST}} & \multirow{1}{*}{\textbf{CelebA}} \\
\midrule
\multirow{3}{*}{\begin{tabular}[c]{@{}c@{}}
  PRISM  \end{tabular}} 
                        & \textbf{FID $\downarrow$} & 34.2038 & 67.1648 & 39.7997 \\
                        & \textbf{P\&R $\uparrow$} & 0.4386 / 0.4236 & 0.4967 / 0.1231 & 0.6294 / 0.0713 \\
                        & \textbf{D\&C $\uparrow$} & 0.1734 / 0.1597 & 0.2748 / 0.1681 & 0.4565 / 0.2967 \\
\midrule
\multirow{3}{*}{\begin{tabular}[c]{@{}c@{}}
  PRISM (vanilla) \end{tabular}} 
                        & \textbf{FID $\downarrow$} & 5.8238 & 5.5004 & 19.1512  \\
                        & \textbf{P\&R $\uparrow$} & 0.6913 / 0.851 & 0.6985 / 0.8534 & 0.6621 / 0.3895  \\
                        & \textbf{D\&C $\uparrow$} & 0.4689 / 0.679 & 0.4864 / 0.6965 & 0.5348 / 0.5947 \\
\bottomrule
\end{tabular}
}
\label{tab:vanilla}
\end{center}
\end{table}

In~\Tref{tab:vanilla}, we report the results of models trained under FL vs. vanilla (centralized data) setups. \prism{} shows performance degradation due to privacy, data heterogeneity, and communication overhead in FL settings. However, centralized training is not applicable in scenarios where distributed samples cannot be shared, which is the primary focus of our work. Therefore, our original submission did not include this comparison, consistent with existing FL research.

\section{Additional results}
\label{appendix:additionalresults}
\subsection{Different random initialization or seed}
\begin{table}[h]
\caption{Quantitative performance of~\prism{} across different random seeds on the MNIST dataset.}
\begin{center}
\resizebox{0.7\columnwidth}{!}{
\begin{tabular}{ccccc}
\toprule
\textbf{IID, DP} & \textbf{seed} & \textbf{FID} & \textbf{P\& R} & \textbf{D\& C} \\ 
\midrule
  \multirow{3}{*}{\begin{tabular}[c]{@{}c@{}} \prism{} \end{tabular}} &  30 & 27.3017 & 0.4377 / 0.5576 & 0.1738 / 0.1982 \\
  & 123 & 26.8707 & 0.4342 / 0.5399 & 0.174 / 0.2047 \\
  & 1234 & 26.9664 & 0.4417 / 0.5842 & 0.1785 / 0.1997 \\
\midrule
\midrule
\textbf{Non-IID, DP} & \textbf{seed} & \textbf{FID} & \textbf{P\& R} & \textbf{D\& C} \\ 
\midrule
    \multirow{3}{*}{\begin{tabular}[c]{@{}c@{}} \prism{} \end{tabular}} &  30 & 34.2048 & 0.4386 / 0.4236 & 0.1734 / 0.1597 \\
      & 123 & 34.0586 & 0.4372 / 0.3306 & 0.1732 / 0.1618 \\
      & 1234 & 32.9423 & 0.3921 / 0.4182 & 0.1496 / 0.1587 \\
\bottomrule
\end{tabular}
}
\label{tab:randomseed}
\end{center}
\end{table}

\prism{} learns a binary mask over frozen, randomly initialized weights. One might argue that the way model parameters are initialized can significantly impact performance. To validate the robustness of~\prism{}, we present quantitative performance across various random seeds using the MNIST dataset in~\Tref{tab:randomseed}. Note that the random seed 30 in the first row of each table corresponds to the default configurations described in~\Sref{sec:experiments} and all of experimental setup is identical to~\Sref{subsec:iid} and~\Sref{subsec:noniid}.

\subsection{Ablation study on the learning rate for baselines}
\begin{table}[h!]
\caption{Baseline's FID performance for adjusting learning rate using MNIST dataset.}
\begin{center}
\resizebox{0.9\columnwidth}{!}{
\begin{tabular}{cccccc}
\toprule
\textbf{Non-IID, DP} & \textbf{1e-4, 1e-4 (default)} & \textbf{1e-3, 1e-4} & \textbf{1e-5, 1e-4} & \textbf{1e-4, 1e-3} & \textbf{1e-4, 1e-5} \\ 
\midrule
  GS-WGAN & diverge & 108.0657 & diverge & diverge & 96.1892 \\
\midrule
\midrule
\textbf{Non-IID, DP} & \textbf{1e-3, 5e-4 (default)} & \textbf{1e-3, 5e-3} & \textbf{1e-2, 5e-4} & \textbf{1e-4, 5e-4} & \textbf{1e-3, 1e-5} \\ 
\midrule
  Dp-FedAvgGAN & 153.9325 & 206.759 & diverge & 177.3752 & 232.2336 \\
\bottomrule
\end{tabular}
}
\label{tab:learningrate}
\end{center}
\end{table}
In this section, we present the additional experiments to validate the fairness of our entire experiments. In~\Tref{tab:learningrate}, we sweep learning rates for GS-WGAN and DP-FedAvgGAN and  a FID performance on MNIST dataset with Non-IID and privacy-preserving scenario. Since they utilize GANs, there are separate learning rates for the generator and discriminator, and we examined these combinations. In the tables below, the top row represents the discriminator’s lr / generator’s lr, while the subsequent row reports the FID scores. Due to the notorious instability of GAN training caused by the coupled learning dynamics between the generator and discriminator, we observed divergence in some settings. For GS-WGAN, the best performance was achieved with 1e-4 / 1e-5, but it still failed to generate MNIST images properly.

\section{Contextualize the practical scenarios}
\label{appendix:practical_scenarios}
Federated learning (FL) is commonly motivated by real-world scenarios such as healthcare, finance, and mobile devices, where direct data sharing is infeasible due to privacy concerns. In this context, our work does not aim to generate private data directly but instead focuses on enabling stable and effective generative modeling under these constraints. For instance, in healthcare, PRISM allows multiple institutions to collaboratively train generative models without sharing sensitive patient data. The resulting models can facilitate privacy-preserving synthetic data generation, imputation of missing medical records, data augmentation for low-data scenarios, and solving inverse problems such as denoising and reconstruction. Similarly, in mobile devices,~\prism{} enables decentralized training of generative models while preserving user privacy, which can enhance personalized content generation and user-specific data augmentation.

\end{document}